\documentclass[10pt]{article}

\usepackage[T1]{fontenc}
\usepackage[utf8]{inputenc}
\usepackage{amsmath}
\usepackage{newtxtext,newtxmath}
\usepackage[letterpaper,margin=1in]{geometry}
\usepackage{microtype}
\usepackage{graphicx}
\usepackage{booktabs}
\usepackage{multirow}
\usepackage{xcolor}
\usepackage{xspace}
\usepackage{enumitem}
\usepackage[font=small,labelfont=bf,labelsep=period,skip=6pt]{caption}
\usepackage{titlesec}
\usepackage[ruled,vlined,linesnumbered]{algorithm2e}
\usepackage[numbers,sort&compress]{natbib}
\definecolor{linkblue}{RGB}{20,70,160}
\usepackage[colorlinks=true,linkcolor=linkblue,citecolor=linkblue,urlcolor=linkblue]{hyperref}
\usepackage[capitalise,nameinlink]{cleveref}
\titleformat{\section}{\sffamily\bfseries\large}{\thesection}{0.6em}{}
\titleformat{\subsection}{\sffamily\bfseries\normalsize}{\thesubsection}{0.6em}{}
\titleformat{\paragraph}[runin]{\bfseries}{}{}{}
\titlespacing*{\section}{0pt}{14pt plus 3pt minus 2pt}{6pt plus 2pt}
\titlespacing*{\subsection}{0pt}{10pt plus 2pt minus 2pt}{4pt plus 1pt}
\titlespacing*{\paragraph}{0pt}{6pt plus 2pt minus 1pt}{0.6em}
\setlist{itemsep=1pt,topsep=3pt}

\graphicspath{{./}}

\newcommand{\method}{\texorpdfstring{\ensuremath{\nabla}Mimic}{NablaMimic}\xspace}
\newcommand{\engine}{NablaEngine\xspace}

\crefname{algocf}{Algorithm}{Algorithms}
\Crefname{algocf}{Algorithm}{Algorithms}

\hypersetup{
  pdftitle={NablaMimic: Ultra-Efficient Humanoid Control with Differentiable Simulation},
  pdfauthor={Weyl Lu, Yubei Chen},
}

\title{\textbf{\method: Ultra-Efficient Humanoid Control with Differentiable Simulation}}
\author{Weyl Lu \qquad Yubei Chen \\[4pt]
UC Davis%
\thanks{Project page: \url{https://sailab.ai/nablamimic/}. Code: \url{https://github.com/sailab-ai/NablaMimic}
  (\method) and \url{https://github.com/sailab-ai/NablaEngine} (\engine).}}
\date{}

\begin{document}
\maketitle

\begin{abstract}
Robot policy learning is dominated by zeroth-order methods such as PPO, which treat the simulator as a black box and
pay for it with enormous sample counts. First-order policy learning differentiates through the dynamics, but on
contact-rich whole-body control it has rarely worked: its gradients explode. A recent world-model method showed that
derivative information can make humanoid motion tracking far more efficient, using surrogate gradients of a learned
model. We show that exact gradients of the simulator itself do better, once two obstacles are removed. Exact gradients
contain rare unbounded samples, because contact dynamics is not differentiable everywhere and its derivative diverges
near some events; and they are expensive, because the simulator's adjoint has to replay the contact solve.

\textbf{\method} bounds these samples with median-relative per-sample gradient clipping at every step of the
backward pass, and reuses each costly simulated window by multistep regression onto gradient-derived action targets, which bounds every action target independently of the gradient scale.
\textbf{\engine}, a differentiable GPU simulation engine for the Unitree G1, removes the cost: it computes MuJoCo's
full contact dynamics with fused per-world kernels and the adjoint of a whole control step as a single megakernel, with
analytic gradients that match float64 finite differences, $11.8\times$ faster than the available MuJoCo Warp adjoint.

On whole-body tracking of all 40 LAFAN1 clips with one policy on one GPU, \method reaches 0.866, the score the
state-of-the-art world-model method reports after 3\,h, in 5.3 minutes; that method's official code, trained in the
same physics, levels off at 0.838. \method reaches 0.890, the plateau of an open-source world-model tracker trained
in the same physics on the same GPU, in 17 minutes, where that tracker needs 2.6\,h, and levels off at 0.929 with no
clip failed. PPO in BeyondMimic's configuration reaches 0.721 after 6\,h, a level \method passes in 1.5 minutes,
and remains below 0.82 in three longer runs of 13 to 17\,h. SHAC, the standard first-order baseline, stops on exploding gradients
within minutes; with its guard relaxed it reaches 0.820 after 3\,h. A \method policy runs unchanged on a physical
Unitree G1. We release \engine and \method as open-source infrastructure for first-order humanoid policy learning.
\end{abstract}

\section{Introduction}
\label{sec:intro}

\begin{figure}[t]
  \centering
  \includegraphics[width=\textwidth]{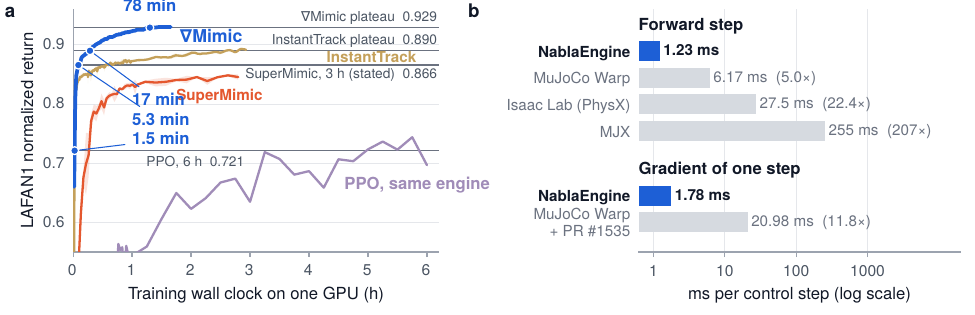}
  \caption{\method trains a high-quality tracking policy in minutes. (a)~LAFAN1 (all 40 clips): normalized return
  against training wall clock (linear; the return axis starts at 0.55) on one RTX PRO 6000, mean $\pm$ sd over 5 seeds for \method and 3 seeds for
  InstantTrack and SuperMimic's official code (both in the same physics; SuperMimic shown to 3\,h), one seed for PPO in
  BeyondMimic's configuration (same engine); every run is scored on MuJoCo Warp. Grey lines mark four
  levels: PPO's after 6\,h (0.721), SuperMimic's reported score after 3\,h (0.866; \citealp[\S4.1, LAFAN1 return curve]{liao2026supermimic}),
  InstantTrack's plateau (0.890) and \method's plateau (0.929). Dots mark where the \method curve first reaches each
  level: after 1.5, 5.3, 17 and 78\,min. (b)~Simulation time per 20\,ms control step for 4096 worlds
  on the same GPU (log scale): forward step, with Isaac Lab on PhysX (its own G1 asset and task defaults) and MJX (the
  same model), and gradient of one step; multiples relative to \engine.}
  \label{fig:hero}
\end{figure}

Motion tracking has become a central task in humanoid learning. A tracking policy imitates reference motions in
physics, so any human motion that can be retargeted to a robot becomes a behavior it can execute. The idea goes back to
DeepMimic and its successors in character animation \citep{peng2018deepmimic,luo2023perpetual,tessler2024maskedmimic},
and has moved to real humanoids \citep{he2024omnih2o,ji2026exbody2,ze2025twist,he2025asap,xie2025kungfubot}. Two
developments made it a foundation rather than a single skill. Large motion datasets of better quality, from AMASS and
LAFAN1 to BONES-SEED, together with robot retargets of them, now cover hours of diverse motion
\citep{mahmood2019amass,harvey2020robust,lafan1retarget2025,bones2026seed}. And a single general tracker trained on
such data can serve as the low-level policy of larger systems: its encoder--decoder structure provides a motion
interface for teleoperation, generation and downstream tasks \citep{chen2025gmt,liao2026beyondmimic,luo2026sonic}.

Almost all of these trackers are trained with zeroth-order policy gradient methods, chiefly PPO
\citep{schulman2017proximal}, which need only rewards from a black-box simulator and pay for it with enormous sample
counts. Massively parallel GPU simulation \citep{makoviychuk2021isaac,mjx2024,mujocowarp2025} made the samples cheap,
yet a whole-body tracking policy still takes hours to days of GPU time, and this cost bounds how far the approach can
scale.

First-order policy learning differentiates the return through the dynamics, so every simulated step contributes an
update direction instead of a scalar reward \citep{freeman2021brax,degrave2019diffphys,xu2022shac,suh2022dodiff}. On
contact-rich whole-body control this promise has rarely held: gradients through stiff contact dynamics explode
\citep{metz2021gradients,suh2022dodiff}. SHAC \citep{xu2022shac}, the standard first-order method, is no exception. On
whole-body tracking with the Unitree G1, every SHAC run hits exploding gradients and stops within minutes
(\cref{sec:explode}).

Policy learning through the derivatives of a learned dynamics model has a long history, from stochastic value
gradients and model-augmented actor-critic to Dreamer and to tracking controllers for physics-based characters
\citep{heess2015svg,clavera2020maac,hafner2020dreamer,fussell2021supertrack,yao2022controlvae,lei2025scalable}.
SuperMimic \citep{liao2026supermimic} recently brought this approach to humanoid whole-body tracking at scale: it
differentiates through a learned world model instead of a simulator and trains a tracking policy far faster than PPO.
We read SuperMimic's result differently: not as evidence for world models, but as evidence for derivatives. The gain
comes from using the dynamics' derivatives at all; the world model is the price SuperMimic pays to make those
derivatives bounded and cheap. This raises the question of what that price buys, and whether it has to be paid. A world model supplies \emph{surrogate gradients}: smooth, bounded and cheap, but derivatives of a
different system. A simulator supplies \emph{exact gradients}: derivatives of the simulated system itself, but with
rare unbounded samples, because contact dynamics is not differentiable everywhere and its derivative diverges near some
events, and expensive, because the simulator's adjoint has to replay the contact solve (\cref{sec:gradients}).

The question is therefore not whether exact gradients beat surrogate ones as they are, but what it takes to use them.
This paper removes both obstacles. \textbf{\method} handles the samples: it bounds the rare extreme gradients per
sample and turns the rest into action targets whose step size is fixed in physical units, so that no single sample can
take over an update (\cref{sec:method}). \textbf{\engine} handles the cost: a differentiable GPU engine specialized to the
G1, with the adjoint of a whole control step as a single megakernel, whose exact gradient costs less than an extra
forward step (\cref{sec:engine}).

On all 40 LAFAN1 clips, with one policy and one GPU, \method reaches 0.866, the score SuperMimic reports after 3\,h,
in 5.3 minutes. SuperMimic's official code, trained in the same physics on the same GPU, levels off at 0.838.
InstantTrack \citep{dao2026instanttrack}, an open-source world-model tracker trained in the same physics on the same GPU, levels
off at 0.890; \method reaches that level in 17 minutes, where InstantTrack needs 2.6\,h, about 9 times as long.
\method itself levels off at 0.929, which it reaches after 78 minutes, with no seed falling on any clip from 57 minutes
on (\cref{fig:hero}). PPO in BeyondMimic's configuration on the same engine reaches 0.721 after 6\,h, a level \method
passes in 1.5 minutes. A policy trained with \method plays training clips on a physical Unitree G1 without any change
(\cref{sec:real}). The engine behind this simulates 18.5\,h of robot motion per
second of wall clock (\cref{fig:hero}b). Motion tracking suits exact gradients, since its reward is dense and
points toward the reference almost everywhere; for sparse rewards, a longer gradient window trades reliance on the
critic against accumulated extreme gradients (\cref{sec:when}).

\paragraph{Contributions.}
\begin{enumerate}
  \item A paired measurement of exact and world-model gradients along the same trajectories: the two agree on the bulk
  of every batch and part ways only in its largest samples, where the exact derivative can be unbounded. This locates the
  failure of prior first-order methods on whole-body control in single samples, not in the gradient as a whole.
  \item \textbf{\method}, a first-order policy learning method that is stable on exact contact gradients at the scale
  of thousands of parallel environments, with each component ablated.
  \item \textbf{\engine}, an open-source differentiable simulation engine for the Unitree G1 whose analytic gradients
  match float64 finite differences.
  \item Humanoid motion tracking beyond the world-model state of the art without a world model, on one GPU: on LAFAN1,
  SuperMimic's 3-hour score in 5.3 minutes, InstantTrack's plateau in 17 minutes instead of 2.6\,h, and a plateau of
  0.929, against 0.838 for SuperMimic's official code in the same physics.
\end{enumerate}
\method and \engine are open source; the links are on the first page.

\section{Related work}
\label{sec:related}

\paragraph{Zeroth-order motion tracking.} Physics-based motion imitation with RL was established by DeepMimic
\citep{peng2018deepmimic}, and universal trackers such as PHC \citep{luo2023perpetual} imitate thousands of clips from
datasets such as AMASS \citep{mahmood2019amass} with one policy. On real humanoids, H2O and OmniH2O
\citep{he2024learning,he2024omnih2o}, ASAP \citep{he2025asap}, GMT \citep{chen2025gmt} and others
\citep{xie2025kungfubot,yin2025unitracker} track retargeted human motion on the Unitree G1. BeyondMimic
\citep{liao2026beyondmimic} tracks agile LAFAN1 motions \citep{harvey2020robust,lafan1retarget2025} with one compact
PPO formulation and shared hyperparameters, and SONIC \citep{luo2026sonic} scales PPO tracking to 700\,h of motion
capture and 21k GPU-hours (128 GPUs for 7 days). These trackers use zeroth-order methods, chiefly PPO
\citep{schulman2017proximal}. We use BeyondMimic's PPO configuration as our zeroth-order baseline.

\paragraph{First-order policy learning in differentiable simulators.} Full-horizon BPTT through differentiable
simulators \citep{degrave2019diffphys,hu2020difftaichi,freeman2021brax} suffers from chaotic, exploding gradients
\citep{metz2021gradients}, and gradients through stiff contact can be biased or high-variance
\citep{suh2022dodiff,suh2022bundled,onoda2026doesdo}. PODS \citep{mora2021pods} regresses the policy onto
gradient-improved actions. SHAC \citep{xu2022shac} introduced short windows \citep{williams1990tbptt} with a
TD($\lambda$) critic; successors adapt the horizon (AHAC; \citealp{georgiev2024ahac}), add maximum entropy (SAPO;
\citealp{xing2025sapo}), combine analytic gradients with PPO (GI-PPO, RPO; \citealp{son2023gippo,zhong2026rpo}) or
decouple rendering from the gradient (D.Va; \citealp{you2025dva}). On legged robots, policies transferred to hardware
were trained with smoothed foot contact \citep{schwarke2025deployable,bagajo2024diffsim2real}, a simplified surrogate
for the backward pass \citep{song2024quadruped}, or an exact forward pass with a smoothed contact sensitivity in the
backward pass \citep{kim2026diffcontactlegged}. DiffMimic \citep{ren2023diffmimic} tracks single clips in Brax without
a critic. BCG \citep{aditya2026bcg} combines SHAC and an ADD imitation reward \citep{zhang2025physics} on the
smoothed-contact simulator of \citeauthor{schwarke2025deployable} with contact-local randomized bundling, and trains G1
trackers for four LAFAN1 clips with 64 environments. These works use 32--1024 environments; we use 2048 BPTT windows
per update on full contact dynamics.

\paragraph{Gradients through learned models.} SVG, MAAC and Dreamer optimize policies through learned dynamics
\citep{heess2015svg,clavera2020maac,hafner2020dreamer}, and MoSim \citep{hao2025mosim} predicts long horizons of
raw physical state with a learned motion simulator. PWM \citep{georgiev2025pwm} and \citet{ma2024awm} report that
learned models can give smoother optimization landscapes than the true dynamics, and \citet{amigo2025decoupled} roll
out in the simulator and backpropagate through a learned model. For physics-based characters, SuperTrack
\citep{fussell2021supertrack}, ControlVAE \citep{yao2022controlvae} and \citet{lei2025scalable} train tracking
controllers through learned dynamics. SuperMimic \citep{liao2026supermimic} trains one world model online with the
policy and reports the state of the art on LAFAN1 and BONES-SEED \citep{bones2026seed}. Their gradients are those of a
learned surrogate; ours are of the simulator itself.

\paragraph{Differentiable and GPU simulators.} GPU simulation
\citep{makoviychuk2021isaac,rudin2021walkinminutes,mittal2025isaaclab,zakka2025playground,zakka2026mjlab} made samples
cheap for zeroth-order methods. Among differentiable engines, DiffTaichi and Warp generate adjoint kernels
\citep{hu2020difftaichi,macklin2022warp}, and dflex \citep{xu2021dflex} and Rewarped \citep{xing2025sapo} use penalty
contacts. Brax and MJX \citep{freeman2021brax,mjx2024} use JAX autodiff; MJX is forward mode only on our model
\citep{mjx_issue2259}. Nimble \citep{werling2021nimble} differentiates LCP contact analytically, and Dojo, Simple and
Ostrich \citep{howell2022dojo,lelidec2024simple,kucera2026ostrich} apply the implicit function theorem at the contact
solution. DiffMJX \citep{paulus2026diffmjx} studies gradients of MuJoCo's soft contacts
\citep{todorov2012mujoco,todorov2014convex}. Analytic rigid-body derivatives \citep{carpentier2018derivatives} and
articulated-body adjoints \citep{qiao2021articulated} avoid generic autodiff. MuJoCo Warp \citep{mujocowarp2025} is not
differentiable in any release; an adjoint exists in an open pull request \citep{mjwarp_pr1535}. Newton
\citep{newton2025} and Genesis \citep{genesis2024} have differentiable modes, which hold the contact normal fixed
\citep{newton_pr2164} or leave some collision operations without gradients, respectively. Kernel fusion for RL
simulation \citep{gleeson2022kernelfusion,shacklett2023madrona} and the run-to-run non-determinism of GPU simulators
\citep{zhang2026gpusimbench} have been studied before. \engine gives exact, deterministic adjoints of MuJoCo's dynamics
for one robot at high throughput.

\paragraph{Clipping and regression onto action targets.} Norm clipping of BPTT gradients goes back to \citet{pascanu2013difficulty}.
Per-sample clipping is standard in private SGD \citep{abadi2016dpsgd}, with median-relative \citep{andrew2021adaptiveclip}
and quantile thresholds \citep{merad2024quantileclip}; clipping helps under heavy-tailed noise and gradient-dependent
smoothness \citep{simsekli2019tailindex,zhang2020whyclipping} but is biased in general \citep{koloskova2023clipping}.
We apply it to the action cotangents inside BPTT, as a heuristic outside these guarantees. Regressing a policy onto
improved actions follows trajectory-optimization teachers \citep{mordatch2014combining,levine2013gps}, MPO
\citep{abdolmaleki2018mpo}, AWR \citep{peng2019awr} and target-based surrogates
\citep{vaswani2022fmapg,lavington2023sso}. We keep PODS's target regression and add what whole-body contact at thousands of parallel environments requires: per-sample median-relative clipping inside BPTT, without which our training diverges or stalls, and a step size fixed in radians by the batch RMS of the clipped gradients (\cref{sec:clip,sec:targets}).

\section{Background}
\label{sec:background}

\paragraph{Task.} Whole-body motion tracking on the Unitree G1 (29 actuated joints), mjlab model \citep{zakka2026mjlab}. Control runs at
50\,Hz with 4 physics substeps of 5\,ms. The policy observes a 10-frame history of noisy proprioception (joint
positions and velocities, base angular velocity, gravity direction, previous action), the next 10 reference frames,
and the orientation of the reference anchor body (torso) relative to the robot's, which needs the IMU but no global
position. The reward is BeyondMimic's tracking reward in SuperMimic's setting
\citep{liao2026beyondmimic,liao2026supermimic}: exponential kernels on the position, orientation and velocities of 14
bodies and on the anchor pose, with action-rate and joint-limit penalties. An episode terminates when the root height
error exceeds 0.25\,m or the tilt error exceeds 0.8. \Cref{app:task} gives the full definitions.

\paragraph{First-order policy learning.} The return of a window of $H$ steps starting from state $s_0$ is
\begin{equation}
  J(\theta) = \sum_{t<H} \gamma^t\, r(s_t, a_t) + \gamma^H V_\phi(s_H),
  \qquad a_t = \pi_\theta(o_t),\quad s_{t+1} = f(s_t, a_t).
  \label{eq:window-return}
\end{equation}
When $f$ is differentiable, $\nabla_\theta J$ follows by backpropagation through time, a pathwise gradient. Unlike the likelihood-ratio gradient of PPO, it uses the derivatives of $f$, so its quality depends on them
(\cref{sec:gradients}).

\section{Exact and surrogate gradients}
\label{sec:gradients}

A first-order method uses the derivatives of the dynamics, which a zeroth-order method ignores. It is therefore
commonly expected to extract more information from each sample, and to be the better choice whenever these derivatives
are available. The record says otherwise: methods that use the simulator's own derivatives have failed on whole-body
control, while a method that uses the derivatives of a learned imitation of the simulator succeeded. This section
explains the paradox. The derivatives of contact-rich dynamics are unusable as they are, the systems that make them well
behaved describe something other than the target dynamics, and computing them costs time. We discuss these trade-offs
and the challenges they pose; the rest of the paper addresses them.

\paragraph{Two sources of gradient.} To discuss these trade-offs we introduce two terms (a surrogate gradient here
is the gradient of a surrogate dynamics, not of a surrogate loss). We call a policy gradient
\emph{exact} when it contains the analytic derivatives of the system the policy is trained and evaluated in: here, when
the dynamics $f$ in \cref{eq:window-return} is the simulator itself. We call it \emph{surrogate} when $f$ is replaced by
another, more tractable system, so that the gradient contains that system's derivatives. The surrogate can be a learned
dynamics model $\hat f$, as in model-based policy optimization
\citep{heess2015svg,clavera2020maac,hafner2020dreamer,fussell2021supertrack,liao2026supermimic}, or a simplified
differentiable simulator that softens contact, for example by penalty springs of limited stiffness in place of rigid
contact \citep{xu2021dflex,xu2022shac,freeman2021brax}. The two are the ends of an axis rather than a dichotomy: a simulator with softened contact lies between them,
and at the two geometric events where the derivative of the step does not exist, \engine{}'s training gradient holds
the undefined term fixed; there is no exact value to depart from there, and everywhere else the two are bit for bit the
same (\cref{sec:engine}). Below we compare the two ends: the exact gradients of \engine{} and the surrogate gradients of a
learned world model, the surrogate SuperMimic showed to work for humanoid tracking.

Neither exactness nor boundedness is what learning needs. What it needs is \emph{learning signal}: the part of a
gradient that improves the return at the scale of one policy update. Two trade-offs decide how much of it
each source delivers per unit of compute (\cref{tab:tradeoffs}).

\begin{table}[t]
\centering
\small
\caption{Exact and surrogate gradients: bias, unbounded derivatives and cost.}
\label{tab:tradeoffs}
\begin{tabular}{@{}p{0.2\linewidth}p{0.37\linewidth}p{0.37\linewidth}@{}}
\toprule
& Exact gradient (simulator) & Surrogate gradient (world model) \\
\midrule
Bias & no model error: derivative of the simulated system & derivative of a different, learned system \\
Unbounded derivatives & near non-differentiable events & none (a smooth network) \\
Cost per update & high (simulator adjoint) & low (one network pass) \\
\bottomrule
\end{tabular}
\end{table}

\paragraph{Unbounded exact, bounded surrogate.} An exact gradient has no bias, but contact-rich dynamics is not
differentiable everywhere. Where contacts appear or vanish, actuators saturate or the solver's active set changes,
the derivative jumps or is undefined. Near some of these events it is unbounded. Computed faithfully, it can be
infinite. We measured this on the G1 (\cref{app:engine}):
\begin{itemize}
  \item when the feet nearly touch, the derivative of the self-contact normal grows as the inverse of their distance;
    on training windows the gradient grew by up to $4.9\times10^{21}$ through a window, and 32 rows were not finite;
  \item for nearly parallel capsules (crossed feet), one control step amplified the gradient by up to $3.6\times10^8$,
    and repeated steps overflowed to non-finite values in 40 of 172 such states.
\end{itemize}
A surrogate gradient, in contrast, is bounded everywhere. For a world model that is a
multilayer perceptron with weight matrices $W_1, \dots, W_n$ and an activation with Lipschitz constant $c$ (SiLU, as in
SuperMimic, has $c \approx 1.1$), the Jacobian of one predicted step satisfies
\begin{equation}
  \Bigl\lVert \frac{\partial \hat f}{\partial (s, a)} \Bigr\rVert_2 \le L = c^{\,n-1} \prod_{l=1}^{n} \lVert W_l \rVert_2
\end{equation}
at every state, and the action gradient of an $H$-step window is bounded by a constant that depends only on $L$, $H$ and
the reward. The bound can be loose and grows with $H$, but it is uniform: no state makes it infinite. A simulator's step Jacobian has no such
bound. The surrogate gradient, however, describes a different system than the one being simulated, so it is biased.

\paragraph{Same bulk, different tail.} The two gradients agree on the bulk of the batch and part ways only at the
extremes. Each on its own rollouts, they have similar tails. With the policy of a SuperMimic reproduction,
the same objective and 32-step windows, the 99.9th percentile of the per-sample action-gradient norm is 25--37 times
the median on both routes, and the maximum a few hundred to about 1,350 times, with no consistent ordering between
them. A paired comparison along the same trajectory shows where the two differ. We ran the same policy in \engine{} and,
at every recorded state and action, computed the action gradient once with the engine's Jacobian and once with the
world model's Jacobian at that state, everything else identical. The two agree closely on most samples (Spearman
correlation of the norms 0.94--0.95, median cosine 0.90--0.91, at two checkpoints) and almost perfectly on small
gradients (cosine 0.96--0.98 for the smallest 10\%). Agreement falls as the exact gradient grows: for the largest
0.1\% of exact gradients, the world model's gradient has only 0.35--0.43 of the norm (median), a median cosine of
0.39--0.67, and in some samples points the opposite way. The world model does not create large gradients of its own:
where its gradient is largest, the exact one is large as well.

The difference is therefore one of tails, not of bias. Along the same trajectory the world model damps the exact tail
(99.9th percentile 16--21 against 24--31 times the median). And with exact derivatives the unbounded events reappear:
without \engine{}'s rules for them, single samples reach $6.5\times10^{24}$ times the median and some are not finite.
Such samples are rare, but one of them dominates a batch-mean gradient. A few samples carry huge or infinite values
and no learning signal; the rest of the batch is as good as the surrogate's, and unbiased.

\paragraph{More exactness is not more signal.} Making the derivative of force-saturated actuators exact changed about 20\% of
per-state gradients but did not improve learning: $-0.023$ after 1\,h (3 seeds, within two seed standard deviations),
and no difference together with the self-contact rule below or after 3\,h. Training with the plain exact derivative
everywhere, without either of \engine{}'s rules and with MuJoCo Warp's actuator convention, gives the same result as the
rules: 0.867 against 0.869 after 1.5\,h, the same time to 0.82 and the same update rate (residual recipe, 3 seeds).

\paragraph{Bias versus cost.} A world model is cheap to evaluate and to differentiate: one network pass, independent
of contact complexity. SuperMimic reports that its world model runs more than 10$\times$ faster than MuJoCo Warp at the
batch sizes used in practice and 62$\times$ faster at 64k parallel environments, and that its training needs over
1000$\times$ fewer GPU-hours than SONIC's cluster-scale PPO pipeline \citep{liao2026supermimic}. The simulators that
robot learning commonly uses are slower still on this robot. One control step of 4096 G1 worlds takes 6.2\,ms in MuJoCo
Warp, 12.9\,ms in Isaac Lab with Newton and 27.5--54.1\,ms with PhysX, 34.7\,ms in MuJoCo C on a 16-thread CPU and
255\,ms in MJX, all forward only (one RTX PRO 6000; \cref{tab:engine-forward}). Few of them provide a gradient of the
full contact dynamics at all: PhysX does not, MuJoCo C offers finite differences, MJX only forward-mode derivatives on
this model, and MuJoCo Warp only through an open pull request.

Per policy update, one SuperMimic iteration takes about 0.1\,s on an H100 (estimated from its paper, \cref{app:repro}). The same
first-order recipe as ours on the MuJoCo Warp adjoint takes 1.63\,s per update, 16$\times$ more. Even on \engine an
update takes 0.38\,s, still more than the world model. Exact gradients make up for it in the number of updates: \method
in its earlier recipe reaches 0.82 after 1.25k updates, about 1/33 of the roughly 41k iterations SuperMimic needs (estimated from its paper; \cref{app:repro} gives the
estimate and an alternative reading).
This is how we read SuperMimic's result. Per update, training on exact gradients needs far fewer: 1.25k updates
against about 41k. Counted in transitions, \method reaches 0.82 after $3.1\times10^8$ simulated transitions, while
SuperMimic's iterations of $4096\times32$ imagined steps amount to $5.4\times10^9$, or $2.1\times10^9$ under the
alternative reading: a factor of 7 to 17 fewer, each of ours simulated rather than imagined. The final recipe of
\cref{sec:method} reaches 0.82 after 572 updates and $8.1\times10^7$ transitions. Per second, at the cost of the simulators available at the time, the surrogate made faster progress.
The world model won on price.

\paragraph{What exact gradients need.} Both trade-offs point at the same two requirements, and at how to meet them
differently from a world model. A surrogate pays one bias, on every sample, to buy two things at once: bounded
derivatives and a cheap update. We buy them separately. Boundedness costs a bias confined to the norms of the samples that
are cut, with their directions and the rest of the batch left exact; this is the job of \method (\cref{sec:method}). Cheapness costs
engineering, not bias: the adjoint has to come down to where the higher learning signal per update pays off in wall
clock; this is the job of \engine (\cref{sec:engine}). Applied to the gradients of a well-trained world model, \method's components bring no gain
(\cref{sec:main-result}): the bias of a surrogate is not what they remove.

\subsection{Prior first-order methods explode}
\label{sec:explode}

\begin{figure}[t]
  \centering
  \includegraphics[width=\textwidth]{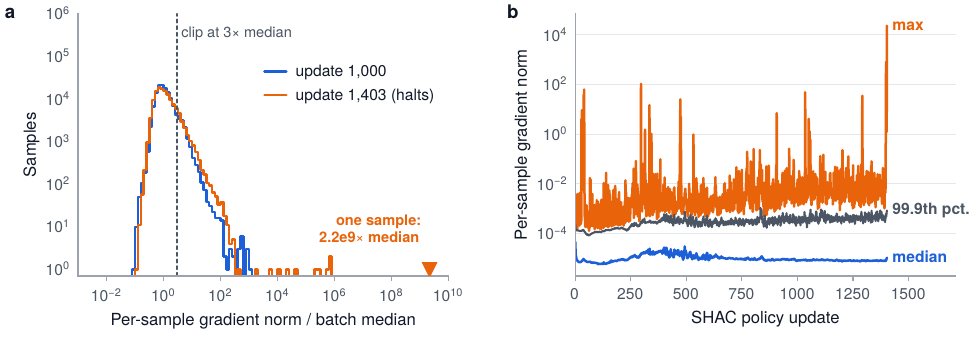}
  \caption{Exploding gradients sit in single samples. Per-sample action-gradient norms of SHAC on exact gradients (one
  seed, raw, before any clipping). (a)~Distribution within one update batch of 131,072 samples, relative to the batch
  median: an ordinary update and the update in which SHAC's guard stops training, where one sample lies at
  $2.2\times10^9$ times the median. (b)~Median, 99.9th percentile and maximum per update: the bulk stays flat, only the
  maximum explodes.}
  \label{fig:grad-tails}
\end{figure}

SHAC \citep{xu2022shac} uses exact gradients the way surrogate-gradient methods use theirs: one global gradient-norm
clip and a guard that stops training when the norm exceeds $10^6$. On whole-body G1 tracking at our scale, every SHAC run
hits exploding gradients and trips this guard: all 3 seeds on \engine within 0.06\,h (746--1568 updates), at a return of
$0.38\pm0.10$, and on MuJoCo Warp within 0.36\,h. The explosion comes from a single sample, not from the batch, and
this is why a global norm clip cannot save it: one sample at $10^9$ times the median either trips the guard or, if
clipped globally, scales every other sample's gradient to nothing. Up to the halt, the median, 99th and 99.9th percentiles of
the per-sample action gradient norms stay flat; in the halting update the largest per-sample gradient is $2.2\times10^9$ to
$1.8\times10^{10}$ times the median in each of three instrumented runs, and this one sample trips the guard. When it
appears is random: the instrumented runs halted after 326 to 8361 updates. \Cref{fig:grad-tails} shows this for one seed.

\subsection{When first-order methods fit}
\label{sec:when}

\paragraph{Dense rewards.} Exact gradients are most useful when the reward is dense and smooth in the state. Motion tracking is such a task: the
reward at every step is a sum of exponential kernels of tracking errors, so its gradient points toward the reference
everywhere except at contact events, and those are what per-sample clipping handles.

\begin{figure}[t]
  \centering
  \includegraphics[width=\textwidth]{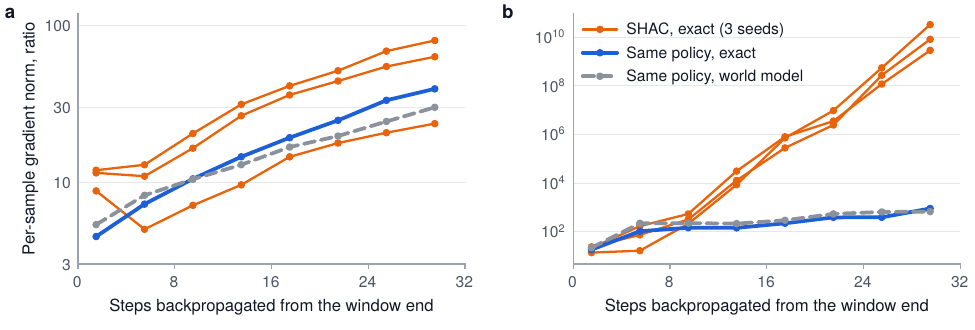}
  \caption{The tail of the per-sample action gradient grows with the number of steps it is backpropagated through,
  while its median stays flat: (a)~99.9th percentile and (b)~maximum, each relative to the median. SHAC: raw exact gradients, 3 seeds, all recorded updates. Same policy: a SuperMimic
  reproduction's policy, the same objective and 32-step windows, exact (\engine{}) and world-model gradients, no
  clipping.}
  \label{fig:tails-depth}
\end{figure}

\paragraph{Horizon trade-off.} With sparse rewards and long-range dependencies the picture changes, and the window
length $H$ sets a second trade-off. A long window lets the gradient reach a distant reward and relies less on the critic
$V_\phi$, whose estimate is biased. But the tail of the gradient grows with every step it is backpropagated through
(\cref{fig:tails-depth}): over 32 steps the 99.9th percentile grows several-fold and, for SHAC's raw gradients, the
maximum by eight to nine orders of magnitude. With the same policy, exact and world-model gradients grow alike, so the
accumulation is a property of long chains of Jacobians, not of contact alone. A world model does not escape this
trade-off; it only starts from a lighter tail.

 A short window keeps the
gradient clean but leaves more to the critic. \Cref{sec:ablation-h} measures this trade-off for tracking. For tasks whose reward lies seconds away, whether differentiability can be exploited depends on
this trade-off, and first-order methods may not pay off.

\section{\method}
\label{sec:method}

\begin{figure}[t]
  \centering
  \includegraphics[width=\textwidth]{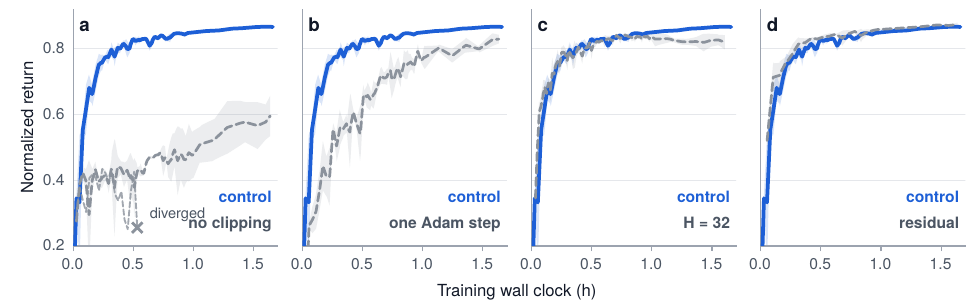}
  \caption{One panel per component of \method: the control (blue) against the recipe with that component changed
  (gray), LAFAN1, mean $\pm$ sd over 3 seeds. (a)~Per-sample clipping: without it one seed diverges and the others
  stay far below. (b)~Multistep regression: one Adam step per window learns more slowly. (c)~Window length: $H = 32$
  plateaus lower. (d)~Residual actions learn faster and plateau higher than absolute actions.}
  \label{fig:ablations}
\end{figure}

We design \method to learn from exact gradients as they are, with their rare extreme samples and a scale that varies
by orders of magnitude (\cref{sec:gradients}). It is a windowed first-order actor-critic built around two components: per-sample clipping and multistep regression onto gradient-derived action targets. The two are a pair.
The action targets fix the size of every update in radians, and clipping keeps any single sample from spending that
budget alone: without clipping, one extreme sample moves its target by about 15\,rad and every other sample's by almost
nothing; with it, no sample moves more than 0.13\,rad (\cref{sec:trust}). Each update simulates a batch of BPTT training windows of $H$ steps on \engine, backpropagates the window return
(\cref{eq:window-return}) through the analytic adjoint, and updates the actor from the resulting action gradients.
\Cref{alg:nablamimic} summarizes one update. Each component has its own subsection and ablation
(\cref{fig:ablations}, \cref{tab:ablations}).

\begin{algorithm}[t]
  \caption{One \method update.}
  \label{alg:nablamimic}
  \DontPrintSemicolon
  \KwIn{actor $\pi_\theta$, critic $V_\phi$, FIFO replay $\mathcal{D}$, collector environments, window length $H$,
  clip factor $\kappa$, step size $\delta$, regression steps $M$ (values in \cref{tab:hyper})}
  \tcp{Data collection (forward only)}
  Step the collector with $\pi_\theta$ and action noise; sample collector resets by return-based bins; push visited
  states to $\mathcal{D}$\;
  \tcp{BPTT training windows on \engine}
  Sample a batch of window start states $s_0$ from $\mathcal{D}$\;
  \For{$t = 0, \dots, H-1$}{
    $a_t \leftarrow \pi_\theta(o_t)$ with action noise\;
    $s_{t+1} \leftarrow f(s_t, a_t)$ (forward pass with tape)\;
    $r_t \leftarrow r(s_t, a_t) + r_{\text{pen}}$ (\cref{eq:term-penalty}); after an episode end, reset the
    environment and continue the window from the reset state\;
  }
  $J \leftarrow \sum_{t<H} \gamma^t r_t + \gamma^H V_\phi(s_H)$ (\cref{eq:window-return})\;
  \tcp{Backward pass with in-backward clipping}
  \For{$t = H-1, \dots, 0$}{
    Propagate the cotangent through step $t$ with the analytic adjoint\;
    Clip each sample's action cotangent to $\kappa$ times the batch median norm (\cref{eq:clip}) before it propagates
    further\;
  }
  Clip the window's action gradients $g_t = \partial J / \partial a_t$ again with \cref{eq:clip}\;
  \tcp{Gradient-derived action targets}
  $y_t \leftarrow a_t + \delta\, g_t / \operatorname{RMS}(g)$ (\cref{eq:target})\;
  \For{$m = 1, \dots, M$}{
    Sample a minibatch of pairs $(o_t, y_t)$; take one step of the separate Adam on
    $\lVert \pi_\theta(o_t) - y_t \rVert^2$\;
  }
  Update the averaged policy $\bar\theta \leftarrow \beta_n \bar\theta + (1 - \beta_n)\,\theta$ (\cref{eq:ema})\;
  \tcp{Critic}
  Fit $V_\phi$ with TD($\lambda$) on pre-transition states\;
\end{algorithm}

\paragraph{Inherited components.} We keep the data path of SHAC and SuperMimic \citep{xu2022shac,liao2026supermimic}
unchanged, so that our comparison isolates the source and handling of the gradient: window start states come from a
replay of states visited by a separate data collector that runs the current policy, the collector's resets are sampled
by return-based bins after BeyondMimic \citep{liao2026beyondmimic}, and a critic $V_\phi$ fit with TD($\lambda$)
bootstraps the window return. Exact gradients make the replay unnecessary: simply continuing each window from where the
previous one ended raises the return by about 0.02 in both recipes (\cref{tab:ablations}). We keep the replay in this
paper so that the comparison with SHAC and SuperMimic isolates the gradient; the simpler data path is also the better
one.

\paragraph{Policy averaging.} The policy we evaluate and deploy is an exponential moving average of the actor's weights;
training uses the live actor. Its decay grows with the number of averaging steps $n$,
\begin{equation}
  \beta_n = \min\!\left(0.999,\ \frac{1 + n}{10 + n}\right),
  \label{eq:ema}
\end{equation}
so early in training the average follows the fast-moving actor closely, and later it averages over about a thousand
updates. Policy averaging, like the residual actions of \cref{sec:residual}, follows InstantTrack
\citep{dao2026instanttrack}.

\paragraph{Fall handling.} A fall ends the episode and bootstraps nothing, so the window return loses the rest of the
episode; the environment then resets and the rest of the window, a new episode, still counts toward $J$, with no
gradient through the reset. SuperMimic's survival weight instead drops everything after the first episode end in a
window \citep{liao2026supermimic}. The critic is fit on the state each transition starts from, and a smooth termination
penalty starts before the termination thresholds ($h$ on the height error, $\theta_{\max}$ on
the tilt), so the gradient sees an approaching fall:
\begin{equation}
    r_{\text{pen}} = -W\,\Delta t \left[ \left(\frac{\operatorname{relu}(d_z - d_{z,0})}{h - d_{z,0}}\right)^{2}
      + \left(\frac{\operatorname{relu}(d_\theta - d_{\theta,0})}{\theta_{\max} - d_{\theta,0}}\right)^{2} \right]
    \label{eq:term-penalty}
  \end{equation}
with $W = 2$, $d_{z,0} = 0.08$, $d_{\theta,0} = 0.10$. Falls enter the gradient only through this smooth margin
penalty; the hard termination itself is never differentiated, and the final policies track all 40 LAFAN1 clips without
a fall (\cref{sec:main-result}). Fall handling is not what separates \method from the baselines:
in the earlier recipe with SuperMimic's masking, removing all of it, the penalty included, changed the return by at most
0.01 at matched wall clock (3 seeds, both action parameterizations).

\paragraph{Effect of averaging and unmasked windows.} Together, policy averaging and continuing the window after a
reset halve the time to SuperMimic's score of 0.82, from 0.34\,h to 0.15\,h, and raise the return at 3\,h from 0.875
to 0.913 (3 seeds each, both with a 500-update learning-rate warmup). Without averaging, removing the mask alone gives $+0.007$ at 1.5\,h and at 3\,h. The component
ablations of this section use the earlier recipe (masked, no averaging).

\paragraph{Networks and optimizer.} The final recipe normalizes every hidden layer of the actor and the
critic with LayerNorm \citep{ba2016layer} before a SiLU activation \citep{elfwing2018sigmoid}, and initializes them
\textmu P-style \citep{yang2022tensor}: input matrices He-normal, hidden matrices with variance $1/\text{fan-in}$,
output matrices zero, with Adam steps of the hidden and output matrices scaled by $256/\text{fan-in}$. The actor's
hidden matrices are trained with Muon \citep{jordan2024muon}, everything else with Adam, in both the actor's update and
the regression onto action targets. The actor's learning rate is tripled for the first 500 updates and returns to its
schedule over the next 300. From update 300 on, every 150 updates the averaged policy runs deterministically from start
frames tiling every clip, and 15\% of the collector's resets go to the 1\,s bins in the 10\,s before each fall it
finds. Muon, the
early learning-rate boost and these evaluation-guided resets follow BlinkMimic \citep{blinkmimic2026}. Each update
uses 2048 windows, half of the earlier recipe's 4096, with window action noise 0.1, and the network matrix
products run in bfloat16 while the physics stays in float32 (\cref{tab:hyper}). With these changes the return after 0.5\,h rises
from 0.872 to 0.908 (\cref{tab:tab_lafan1_paper}). The core components of \method are ablated again on this final
recipe (\cref{tab:abl-final}); the network and optimizer changes are evaluated together, and the remaining ablations
of this section use the earlier recipes.

\begin{table}[t]
\centering
\small
\caption{Component ablations on LAFAN1: one change to the recipe at a time, 3 seeds each, against a 3-seed control at
matched wall clock (about 1.6\,h on one RTX PRO 6000, where the control is still rising and differences are largest;
\cref{app:lafan1} gives the full curves). Change in normalized return for the absolute-action recipe
(control 0.862), on which the ablations were first run (\cref{fig:ablations}), and for the residual-action recipe of
\method (control 0.871). Full tables in \cref{app:lafan1}.}
\label{tab:ablations}
\resizebox{\linewidth}{!}{%
\begin{tabular}{@{}lcc@{}}
\toprule
Change & Absolute actions & Residual actions (\method) \\
\midrule
No per-sample clipping & $-0.292$ (1 of 3 seeds diverged) & all 3 seeds diverged \\
No multistep regression (one Adam step per window) & $-0.053$ & $-0.041$ \\
$H = 32$ instead of 64 & $-0.040$ & $-0.040$\textsuperscript{a} \\
Residual instead of absolute actions (a parameterization) & $+0.008$ & --- \\
No replay starts (continued windows) & $+0.018$ & $+0.020$ \\
\bottomrule
\end{tabular}}
\par\smallskip\footnotesize\textsuperscript{a}\,Mean of 3 seeds; a fourth seed collapsed and is listed separately in
\cref{tab:tab_ablations_res}.
\end{table}

\subsection{Per-sample median-relative clipping}
\label{sec:clip}

The extreme exact gradients of \cref{sec:gradients} live in individual samples: a few environments near contact
events carry action cotangents orders of magnitude above the rest and dominate a batch-mean gradient. A global norm
clip, as in SHAC, rescales the whole batch and destroys the learning signal of every other sample. \method instead
clips the action cotangent of each sample, the gradient of the window return with respect to the action $a_t$ of one
environment at one step, to $\kappa$ times the batch median norm. This happens at every step of the backward pass,
before the cotangent flows into the policy and from there into earlier states, and again on the window's action
gradients; the cotangent that flows directly through the dynamics from one state to the previous one is left to the
engine's rules for unbounded derivatives (\cref{sec:engine}):
\begin{equation}
  g_i \leftarrow g_i \cdot \min\!\left(1,\ \frac{\kappa \cdot \operatorname{median}_j \lVert g_j \rVert}{\lVert g_i
  \rVert}\right), \qquad \kappa = 3.
  \label{eq:clip}
\end{equation}
The median is the reference because it is the one statistic the explosion never moves: in \cref{fig:grad-tails}b the
median and the 99.9th percentile stay flat through the update that halts SHAC, and only the maximum leaves. A
median-relative threshold therefore follows the gradient scale over training, needs no absolute value to tune, and never
cuts the backward chain. With $\kappa = 3$, clipping rescales 9--21\% of the (step, environment) pairs per update; it
preserves their direction and changes only their norm. The samples that matter are far fewer: those at $10^4$ to $10^6$
times the median, the tail of \cref{fig:grad-tails}a, one of which is enough to take over a batch-mean update. The tail does not go away as training proceeds: in an instrumented \method run, the
largest per-sample action gradient is typically $10^4$ to $10^5$ times the batch median at every stage, and up to
$1.9\times10^6$ times; over about 92,000 updates (3 seeds) no gradient was non-finite. Clipping is not free: the
gradients of the clipped samples are biased. It trades this bias for stability.

Without clipping, one of three seeds diverges after 0.54\,h and the other two reach 0.570, against 0.862 for the
control (\cref{fig:ablations}a). Clipping is what makes exact gradients usable at all. Its threshold matters much less:
with $\kappa = 1$, 10 or 30 every seed trains and ends 0.015 to 0.02 below $\kappa = 3$ after 1.5\,h; without clipping all
three seeds diverge (residual recipe, 3 seeds each; \cref{tab:sensitivity}). The final recipe confirms both points
(\cref{tab:abl-final}): without per-sample clipping every seed collapses within about 100 updates, because a few
samples with very large gradients dominate the RMS of \cref{eq:target} and the step of every other sample shrinks
toward zero; clipping only the window's action gradients, not the cotangents inside the backward pass, trains as well
as clipping at every step (0.866 held from 6.4 against 5.6\,min, and a mean return over updates 21k--24k of
$0.9297\pm0.0025$ against $0.9315\pm0.0015$, 3 seeds each).

\subsection{Multistep regression with gradient-derived action targets}
\label{sec:targets}

Every exact gradient is expensive: it costs a simulated window and its adjoint. A world-model method can afford to use
each imagined window for one gradient step and discard it; we want to extract as much as possible from each simulated
window. Taking several gradient steps on the same window, however, would move the policy beyond the region in which
that gradient describes the return, and the gradient's scale, which varies by orders of magnitude over training and
across the batch, would set the step size. We therefore adopt policy optimization by differentiable simulation (PODS)
\citep{mora2021pods}: the gradient defines targets in action space, and the policy is fit to them by several steps of
regression. \method turns the clipped action gradient $g_t$ of each window step into a target action
\begin{equation}
  y_t = a_t + \delta \cdot \frac{g_t}{\operatorname{RMS}(g)},
  \label{eq:target}
\end{equation}
so that the RMS setpoint change over the batch is $\delta$, in radians and decaying with the learning-rate schedule, and regresses the policy onto the targets with $M$
minibatch steps under a separate Adam optimizer \citep{kingma2015adam} (settings in \cref{tab:hyper}). The multistep
regression reuses each simulated window several times, while the regression stops pulling once the policy reaches $y$:
the bounded targets regularize each update in physical units, without line search or KL term (\cref{sec:trust}). With one regression step and a matched learning rate the update reduces to a direct gradient
step [\citealp[Eq.~10]{mora2021pods}; \citealp[Thm.~1]{you2025dva}]; the difference lies in the multistep regression,
the physical step size and the separate optimizer state.

Replacing the multistep regression by one Adam step per window on the raw gradient lowers the return to 0.809
($-0.053$) and slows learning throughout (\cref{fig:ablations}b). In the final recipe, a single regression step per
update reaches 0.866 about 7 minutes later and 0.890 on two of three seeds within 54 minutes, and replacing the RMS
normalization of \cref{eq:target} by a fixed gain reaches 0.866 about 6 minutes later (\cref{tab:abl-final}).

\subsection{Clipping and multistep regression together: bounded targets}
\label{sec:trust}

The two components bound the update in complementary ways. Let $g_i$ be the clipped action gradient of sample $i$ (one
step of one window, $d = 29$ joints), $m$ the batch median of $\lVert g_i \rVert$ and $N$ the number of samples. The
target of sample $i$ moves its setpoints by $\delta\, g_i / \operatorname{RMS}(g)$ (\cref{eq:target}), where the RMS runs
over all samples and joints, so the per-joint RMS change of sample $i$ is
\begin{equation}
  \frac{\delta\, \lVert g_i \rVert}{\sqrt{d}\, \operatorname{RMS}(g)} .
\end{equation}
With clipping, $\lVert g_i \rVert \le \kappa m$, and at least half of the samples have $\lVert g_i \rVert \ge m$, so
$\operatorname{RMS}(g) \ge m / \sqrt{2d}$. Every sample's per-joint RMS change is then at most $\sqrt{2}\,\kappa\,\delta$
(0.13\,rad with our settings), whatever the scale of the gradients: each sample's target stays within a bounded
distance of its current action. The bound holds for the targets the policy is regressed onto, and the regression
inherits it as a regularizer. In practice the policy moves far less: after the regression steps of an update, the typical
environment's setpoints change by 0.002 to 0.01\,rad (per-joint RMS), and the largest change in any environment over a
whole training run is 0.11\,rad. Without clipping, a single sample with norm $G$ far above the others gives
$\operatorname{RMS}(g) \approx G / \sqrt{Nd}$; that sample's target moves by about $\delta \sqrt{N}$ (about 15\,rad for
our batch size), and the targets of all other samples barely move. The step-size budget of the whole batch goes to
one sample, and the learning signal of the rest is lost. The multistep regression fixes the size of the step; clipping keeps any
single sample from taking it over.

\subsection{Window length}
\label{sec:ablation-h}

The window length sets the horizon trade-off of \cref{sec:when}: longer windows rely less on the critic, shorter ones
accumulate fewer extreme gradients (\cref{fig:tails-depth}) and cost less per update. Halving the window to $H = 32$
lowers the return to 0.821 ($-0.040$) at matched wall clock and by 0.046 at matched updates (3 seeds;
\cref{fig:ablations}c). The sweep on the residual recipe shows both ends of the trade-off (return after 1.5\,h, 3 seeds
each; \cref{tab:sensitivity}). Short windows fail through the critic: at $H = 16$ every seed diverges within minutes,
the critic's value running away, and at $H = 32$ one of four seeds collapses the same way while the others reach
0.815--0.841. Long windows are stable but slow: $H = 128$ doubles the backward pass of each update, completes half as many
updates and reaches 0.839. $H = 64$ (0.869, no collapse in any run) lies between the two.

\subsection{Residual actions}
\label{sec:residual}

The policy outputs an offset to the reference joint pose; the PD setpoint is reference plus offset. An untrained
policy then already follows the reference pose, so learning refines a correction instead of reproducing the motion
from scratch. The action-rate
penalty is applied to the commanded setpoints, so the reward, and hence the task, is identical to predicting absolute
setpoints; only the parameterization changes. Residual actions learn faster (\cref{fig:ablations}d) and end higher:
in the earlier recipe (masked, no averaging), they raise the return at matched wall clock (1.6\,h) from 0.862 to
0.870 (3 seeds each; \cref{tab:ablations}).

\section{\engine}
\label{sec:engine}

\engine is a differentiable simulation engine for the Unitree G1 that makes exact gradients cheap. It computes the
dynamics of MuJoCo \citep{todorov2012mujoco} for this robot, including the full contact model with all 502 contact pairs and
the Newton constraint solver, and their analytic derivatives. It is specialized to one robot by design: fixing the
kinematic tree, the contact pairs and the actuator layout at compile time is what lets one warp hold a world's step in
registers and shared memory, and what makes the adjoint cheaper than another forward step. Its design rests on three
ideas.

\paragraph{One world per warp, a megakernel for the adjoint.} \engine specializes the step to one robot and maps each
world to one warp. Within a kernel, the world's state stays in shared memory and registers. The forward step runs as two
fused kernels per substep, 8 physics kernels per 20\,ms control step. The adjoint of a whole control step is a single
megakernel: one warp per world reads the tape and runs the derivatives of all four substeps in reverse without leaving
the kernel. (The forward step is split into two kernels per substep for occupancy; \cref{app:engine}.)

\paragraph{A hand-derived adjoint.} The backward pass is derived by hand for every stage: kinematics, collision, the
constraint solve and integration. The constraint solve is differentiated through its optimality conditions, not
through the solver's iterations, with the discrete decisions (contact set, active constraints, actuator clamps) held
fixed. The forward pass writes a compact tape, and the adjoint replays it in reverse. The result is the derivative of
the discrete step: it matches float64 finite differences of MuJoCo to $10^{-5}$ in median relative error, and the
forward step matches float64 MuJoCo to $8.5\times10^{-7}$ (\cref{app:accuracy}); the forward pass is bitwise deterministic.

\paragraph{Bounded where the derivative is not.} Near two non-differentiable events of the G1's geometry the exact
derivative is unbounded (\cref{sec:gradients}): self-contact at vanishing distance and nearly parallel capsules. For
training, \engine freezes the corresponding geometric derivatives below a small threshold (1\,mm, about $2^\circ$), and on
force-saturated actuators it keeps the actuator damping term, which the exact derivative drops; making the latter exact
does not improve learning (\cref{sec:gradients}). Away from these events the training gradient equals the exact
gradient bit for bit. These rules are part of the method, not an approximation of it: they define the gradient where
the exact derivative is undefined or infinite and leave it untouched elsewhere. \Cref{app:engine} gives the
details.

\paragraph{Speed.} \Cref{tab:engine-speed} compares \engine with MuJoCo Warp \citep{mujocowarp2025}, the GPU
implementation of the same dynamics. A forward step is $5.0\times$ faster on an RTX PRO 6000; a step with its gradient
is $11.8\times$ faster than the open pull request that adds an adjoint to MuJoCo Warp \citep{mjwarp_pr1535}, the only adjoint of MuJoCo's full
contact dynamics on a GPU; MJX differentiates this model in forward mode only, and the other engines of
\cref{tab:engine-forward} provide no gradient of it. On
\engine{} the gradient costs less than another forward step: a step with its gradient takes 1.78\,ms against 1.23\,ms
for the step alone, and the backward pass of the entire adjoint, contact solve included, takes 0.56\,ms when timed as a
separate call (\cref{fig:engine-split}). One RTX PRO 6000 simulates 3.33\,M control steps per second, about 18.5\,h of
robot motion per second of wall clock.
Other engines and GPUs are in \cref{app:engine}.

\begin{table}[t]
  \centering
  \small
  \caption{Time per 20\,ms control step for 4096 G1 worlds, ms (mean of the medians on two held-out state sets; V100: one set).}
  \label{tab:engine-speed}
  \begin{tabular}{@{}lrrrr@{}}
    \toprule
    & RTX PRO 6000 & H100 & RTX 3090 & V100 \\
    \midrule
    \multicolumn{5}{@{}l}{\emph{Forward step}} \\
    \engine & \textbf{1.23} & \textbf{1.72} & \textbf{3.42} & \textbf{5.79} \\
    MuJoCo Warp & 6.17 & 8.69 & 12.88 & 18.4 \\
    \midrule
    \multicolumn{5}{@{}l}{\emph{Step and its gradient}} \\
    \engine & \textbf{1.78} & \textbf{2.74} & \textbf{5.26} & \textbf{10.6} \\
    MuJoCo Warp with adjoint pull request & 20.98 & 25.13 & 42.74 & 63.2 \\
    \bottomrule
  \end{tabular}
\end{table}

\section{Experiments}
\label{sec:experiments}

Component ablations sit next to each component in \cref{sec:method}; this section reports the main comparison against
the other gradient sources and separates the engine's contribution from the algorithm's.

\subsection{Setup}
\label{sec:setup}

We train one policy on all 40 LAFAN1 clips \citep{harvey2020robust} retargeted to the G1 \citep{lafan1retarget2025}
(264,705 frames, 2.45\,h of motion). Each clip is evaluated once from its first frame with a deterministic actor and no
randomization or noise. The score is the full-clip normalized return as defined by
\citet[\S4]{liao2026supermimic}: the tracking reward minus the action-rate and joint-limit penalties, summed over the
whole clip and divided by its maximum, so that a fall forfeits the rest of the clip; it is averaged over the 40 clips,
and its linear-velocity term compares reference and robot at the same point of each body, so that a perfect tracker
scores 1 (\cref{app:eval}). Every LAFAN1 run is scored with this protocol on MuJoCo Warp with 10 solver iterations,
the simulator none of the learning methods trains in, so that all methods are measured by the same independent
plant. Scores on MuJoCo Warp are not bit-reproducible: re-scoring one checkpoint varies by up to about 0.004 without
any fall, so differences below about 0.004, or a single fallen clip, are not differences between methods.
Policies transfer between the two simulators: the final \method checkpoints score $0.930\pm0.004$ on MuJoCo Warp and
$0.930\pm0.003$ on \engine{}, after 3\,h of training the earlier recipes score the same on both (per seed within
0.001), and at 6{,}000 updates a policy scores at most about 0.01 higher in the simulator it was trained in, whichever
that is (\cref{app:sim2sim}).
Wall clock counts training only, without start-up compilation or evaluation. The plateau of a run is the mean return
over its last 30 minutes of training, averaged over seeds; the time to a level is the first crossing of the seed-mean
curve, counted from the first training update. Each run uses one RTX PRO 6000 unless stated, and each configuration
three seeds (five for the main \method runs; mean $\pm$ sd). Seeds that diverge or collapse are reported
separately and not averaged. A seed sets the network initialization and all random draws (action and observation
noise, resets and randomization); the engine itself is deterministic.

The baselines cover the alternatives to our gradient source. SuperMimic \citep{liao2026supermimic}, the surrogate-gradient
state of the art, enters twice: with the numbers its paper states (one H100, trained without self-collisions), and with
its official code trained in our physics, which includes the robot's self-collisions: the released commit, with
proprioceptive actor input, a 10-frame observation history and all 40 clips, 3 seeds of about 3.2\,h, otherwise as
released, scored by our protocol (\cref{app:supermimic}). The public reproduction we use for the gradient analysis of
\cref{sec:gradients} is described in \cref{app:repro} and is not a baseline. PPO \citep{schulman2017proximal} on \engine{} is the zeroth-order reference, once in BeyondMimic's configuration
\citep{liao2026beyondmimic} and once with SuperMimic's actor and reset sampling. SHAC \citep{xu2022shac} is the
first-order reference (\cref{sec:explode}). InstantTrack \citep{dao2026instanttrack}, an open-source implementation of SuperMimic's method (a learned world
model and backpropagation through it) with policy averaging, residual actions and an optimized GPU training loop, is
a much stronger reproduction of SuperMimic than our own (\cref{app:repro}), and the strongest public one we know of.
It is therefore the surrogate-gradient baseline we run: its latest release, version 1.4, with its released training code and
settings and with \engine{} as its simulator on the shared G1 model, on the same GPU as \method, 3 seeds of 3\,h each, with its averaged policy
scored by our protocol. Our physics differs from the plant InstantTrack was developed in, which has no
self-collisions; there the same code reaches $0.900\pm0.001$ (\cref{app:instanttrack}).

\subsection{Main result: minutes instead of hours}
\label{sec:main-result}

We compare the methods by the time they need to reach four levels (\cref{tab:levels}): 0.721, the level PPO in
BeyondMimic's configuration reaches after 6\,h; 0.866, the score of SuperMimic's LAFAN1
checkpoint after 3\,h \citep[\S4.1, its LAFAN1 return curve]{liao2026supermimic}; 0.890, the plateau of InstantTrack; and 0.929, the plateau of \method. \method reaches
SuperMimic's 3-hour score in 5.3\,min (per seed 4.8--6.0\,min; \cref{fig:hero}a) and InstantTrack's plateau in 17\,min
(per seed 13--19\,min), where InstantTrack itself needs 2.6\,h, about 9 times as long; InstantTrack reaches
SuperMimic's score after 33\,min. Both pass the PPO level in 1.5\,min, while PPO needs 5.0\,h to get there.
SuperMimic also states that it reaches 0.82 within 1.2\,h on one H100; \method reaches 0.82 in 2.5 minutes, and its
absolute-action version, in the earlier recipe, in 0.40\,h on an H100 (\cref{app:lafan1}). Our wall clock counts
training only.

SuperMimic's official code, trained in the same physics, reaches 0.721 after 16.1\,min and 0.82 after 45.5\,min (on
the RTX PRO 6000 clock), peaks at 0.852 and levels off at $0.838\pm0.021$; none of its
seeds reaches 0.866 within its 3.2\,h budget. Its paper reports 0.866 after 3\,h in a plant without self-collisions;
in ours, its plateau is a level \method passes after 2.9\,min. The same ordering holds for the quality of
tracking: at the final checkpoints, \method completes every clip on every seed with a mean body position error of
2.16\,cm, against 2.57\,cm for InstantTrack and 3.45\,cm with 97.5\% of clips completed for SuperMimic's official
code (\cref{tab:tracking-metrics}), and on the 30 clips longer than 200\,s, played without the root position in the
observation, the root drifts on average 0.15\,m from the reference, half as far as InstantTrack's and an eighth of
SuperMimic's (\cref{tab:drift}).

\method keeps improving until it levels off at $0.929\pm0.002$, which the seed-mean curve reaches after 78\,min (per
seed 73--89\,min, four of five seeds); at the end of training, 24k updates (1.7\,h), the return is $0.930\pm0.004$.
All five seeds first complete every clip at update 1{,}250 (5.3\,min), and no seed falls on any clip from 57\,min
(update 13{,}750) on, and the five seeds agree to a standard deviation of 0.004
at the end. No run
diverged; the collapses in \cref{tab:ablations} occur only in
variants that remove clipping or change the window.
Starting every training window from the live rollout state, without the replay of collected start states, speeds training further: in a later run of this variant of the recipe (3 seeds; minutes from the update count at its measured 0.238\,s per update on the RTX PRO 6000), the seed-mean curve passes the 6-hour PPO level after 1.2\,min, SuperMimic's reported score after 4.4\,min and InstantTrack's plateau after 8.3\,min, and two seeds trained to 24k updates on that GPU level off at $0.938\pm0.004$.

The other gradient sources on the same engine are far behind. PPO with BeyondMimic's configuration reaches 0.635 after
3\,h and averages 0.721 over its last 30 minutes before the 6\,h budget ends, a level it first reaches after 298\,min
(best checkpoint 0.744, final checkpoint 0.697), even though \engine{} makes each PPO iteration $2.9\times$ faster than stock MuJoCo Warp (0.123
vs 0.355\,s): cheaper samples do not close the gap, the gradient does. In transitions, \method reaches 0.82 after $8.1\times10^7$;
PPO consumes $1.7\times10^{10}$ in 6\,h without reaching it. Three further seeds of the same configuration, trained
for 48\,h on RTX 3090 GPUs (one seed finished on an A40), cover 13.6--16.8\,h on the RTX PRO 6000 clock (converted by
update count); their seed-mean return is 0.721 after 4.4\,h (per seed $4.0\pm0.3$\,h), 0.771 after 10\,h, and $0.782\pm0.005$ over the last 30 minutes, still rising
slowly and never at 0.82. This matches the independent PPO result in SuperMimic's
paper: PPO-Sim reaches 0.60, 0.70 and 0.72 after 18\,h on 1, 4 and 8 H100s. SHAC stops on its gradient guard at 0.06\,h; with
the guard relaxed so that exploding updates are clipped instead of halting training, it keeps learning, reaches 0.820
at 3\,h, and its seed-mean curve crosses 0.82 after 2.6\,h (per seed $2.1\pm0.3$\,h), more than 50 times later than
\method on the same engine. \Cref{tab:tab_lafan1_paper} gives the returns at fixed wall clock; all curves and the cost
to reach 0.82 are in \cref{app:lafan1}.

\begin{table}[t]
\centering
\small
\caption{LAFAN1, all 40 clips: plateau and minutes of training to four levels, PPO's level after 6\,h (0.721),
SuperMimic's stated score after 3\,h (0.866), InstantTrack's plateau (0.890) and \method's plateau (0.929). Plateau: per
seed the mean return over the last 30\,min of training, then mean $\pm$ sd over seeds. Time to a level: first crossing
of the seed-mean curve, in minutes from the first training update; per-seed ranges in the text. All runs in the same
physics, scored on MuJoCo Warp; one RTX PRO 6000 per run, or its clock by update count where marked. PPO's returns
from 2.5\,h on are means of three evaluations per checkpoint, which differ by up to 0.03.}
\label{tab:levels}
\setlength{\tabcolsep}{5pt}
\begin{tabular}{@{}lccccc@{}}
\toprule
Method & Plateau & To 0.721 & To 0.866 & To 0.890 & To 0.929 \\
\midrule
\method (5 seeds) & \textbf{0.929}\,$\pm$\,0.002 & \textbf{1.5} & \textbf{5.3} & \textbf{17.2} & \textbf{78} \\
InstantTrack v1.4 (3 seeds) & 0.890\,$\pm$\,0.001 & \textbf{1.5} & 33.2 & 155 & not reached \\
SuperMimic, official code (3 seeds)\textsuperscript{a} & 0.838\,$\pm$\,0.021 & 16.1 & not reached & not reached & not reached \\
PPO, BeyondMimic configuration (1 seed, 6\,h) & 0.721 & 298 & not reached & not reached & not reached \\
PPO, BeyondMimic configuration (3 seeds, 13.6--16.8\,h)\textsuperscript{b} & 0.782\,$\pm$\,0.005 & 266 & not reached & not reached & not reached \\
\bottomrule
\end{tabular}
\par\smallskip\raggedright\footnotesize
\textsuperscript{a}\,Two seeds trained on H100; training time converted to the RTX PRO 6000 by update count
(0.315\,s per update); no seed exceeds 0.852. \quad
\textsuperscript{b}\,48\,h of training on RTX 3090 GPUs (one seed finished on an A40), clock converted to the RTX PRO
6000 by update count (0.1227\,s per update): 13.6--16.8\,h; per-seed time to 0.721 $240\pm19$\,min; still rising
slowly at the end.
\end{table}

\paragraph{Is it the gradient or the recipe?} This analysis and the next use InstantTrack version 1.1 and the
center-of-mass measure (\cref{app:eval}). To separate the gradient from the engine and the recipe, we train
InstantTrack on \engine{}: its world model learns \engine{}'s dynamics, and its imagined windows carry the same
action noise ($\sigma=0.1$) as its collected data. Once with its own algorithm and once with \method's per-sample
clipping and action targets on its world-model gradients, 3 seeds each, compared at matched updates on one RTX PRO
6000. InstantTrack reaches 0.82 after 11--12.5\,min and 0.866 after 48--53\,min per seed, and 0.879 at 2\,h. With
\method's components it is no better: 0.82 after 10--18\,min, 0.866 after 51--62\,min, 0.845 at 2\,h. \method on the
same engine reached 0.82 in 2.7\,min and 0.866 in 7.5\,min on that measure. The components do not make the surrogate
gradient fast.

The gradients themselves differ little per step. On the same start states, action noise and objective, InstantTrack's
world-model gradient of a window and \engine{}'s exact gradient have a median cosine of 0.76--0.88 at checkpoints
from 0.25 to 4\,h, and a positive cosine on 98--99\% of the samples. A step of 0.03\,rad RMS along the clipped
world-model direction achieves 95--97\% of the true improvement of the window objective that the same step along the
clipped exact direction achieves after 1\,h, and 87--89\% at 0.25\,h; neither makes more than 0.6\% of the windows
worse. Agreement is highest
for the first steps of a window (median cosine 0.88--0.94 over steps 1--4) and lowest for its last steps (0.70--0.84
over steps 17--32) and in flight phases. Per step, the exact gradient is at least as good as a well-trained world
model's, and it needs no model. Per policy update, the exact adjoint gives the larger improvement; a world-model update
is cheaper because the learned model replaces the differentiable simulator step, which is where its wall-clock
advantage comes from. With a weaker world model the gap is large (\cref{app:wm-grad}).

\begin{table}[t]
\centering
\small
\caption{LAFAN1, all 40 clips: normalized return at equal training wall clock, time to 0.82, the score SuperMimic
reports within 1.2\,h (first crossing of the seed-mean curve), and clips with a fall at the end of training. Ours: one
RTX PRO 6000 per run, mean $\pm$ sd over 3 seeds unless noted (\method: 5 seeds); all train on \engine{} except
SuperMimic's official code, which trains with its own world model in the same physics; all are scored on MuJoCo Warp.
InstantTrack: \cref{tab:levels}.}
\label{tab:tab_lafan1_paper}
\setlength{\tabcolsep}{5pt}
\resizebox{\linewidth}{!}{%
\begin{tabular}{@{}lcccccc@{}}
\toprule
Method & 0.5\,h & 1\,h & 2\,h & 3\,h & To 0.82 & Falls (/40) \\
\midrule
\method & \textbf{0.908}\,$\pm$\,0.002 & \textbf{0.923}\,$\pm$\,0.002 & \multicolumn{2}{c}{\textbf{0.930}\,$\pm$\,0.004 at 1.7\,h\textsuperscript{g}} & \textbf{0.04\,h} & 0.0 \\
\method, earlier recipe\textsuperscript{h} & 0.872\,$\pm$\,0.006 & 0.892\,$\pm$\,0.011 & 0.905\,$\pm$\,0.016 & 0.916\,$\pm$\,0.003 & 0.12\,h & 0.0 \\
\method, absolute actions\textsuperscript{e} & 0.829\,$\pm$\,0.005 & 0.850\,$\pm$\,0.004 & 0.867\,$\pm$\,0.001 & 0.870\,$\pm$\,0.002 & 0.48\,h & 0.0 \\
SuperMimic, official code\textsuperscript{d} & 0.800\,$\pm$\,0.013 & 0.833\,$\pm$\,0.001 & 0.843\,$\pm$\,0.004 & 0.847\,$\pm$\,0.006 & 0.76\,h & 0.9 \\
PPO, BeyondMimic configuration\textsuperscript{a} & 0.454 & 0.544 & 0.623 & 0.635 & not reached & 8 \\
PPO, SuperMimic actor and resets\textsuperscript{a} & 0.428 & 0.477 & 0.580 & 0.643 & not reached & 5 \\
SHAC (changed guards) & 0.727\,$\pm$\,0.017 & 0.792\,$\pm$\,0.010 & 0.810\,$\pm$\,0.002 & 0.820\,$\pm$\,0.006 & 2.60\,h\textsuperscript{c} & 1.7 \\
SHAC (published guard) & \multicolumn{4}{c}{stopped at 0.06\,h by its gradient guard (return 0.384\,$\pm$\,0.102)} & not reached & 22.7 \\
SuperMimic (stated)\textsuperscript{b} & n/s & n/s & n/s & 0.866 & within 1.2\,h & n/s \\
\bottomrule
\end{tabular}}
\par\smallskip\raggedright\footnotesize
\textsuperscript{a}\,One seed, 6\,h budget; falls at 6\,h. \quad
\textsuperscript{b}\,One H100, as stated in \citet[\S4.1]{liao2026supermimic}; n/s: not stated. \quad
\textsuperscript{c}\,Per-seed crossings $2.13\pm0.34$\,h. \quad
\textsuperscript{d}\,Clock converted to the RTX PRO 6000 by update count; falls are means over three evaluations of
the final checkpoint where repeated. \quad
\textsuperscript{e}\,Masked windows, no policy averaging (\cref{sec:method}). \quad
\textsuperscript{g}\,End of training (24k updates); the plateau, the mean over the last 30\,min, is $0.929\pm0.002$. \quad
\textsuperscript{h}\,The earlier main recipe (4096 windows per update, Adam, no normalization layers;
\cref{sec:method}), 3 seeds.
\end{table}

\subsection{Same algorithm, two simulators}
\label{sec:two-simulators}

\begin{figure}[t]
  \centering
  \includegraphics[width=\textwidth]{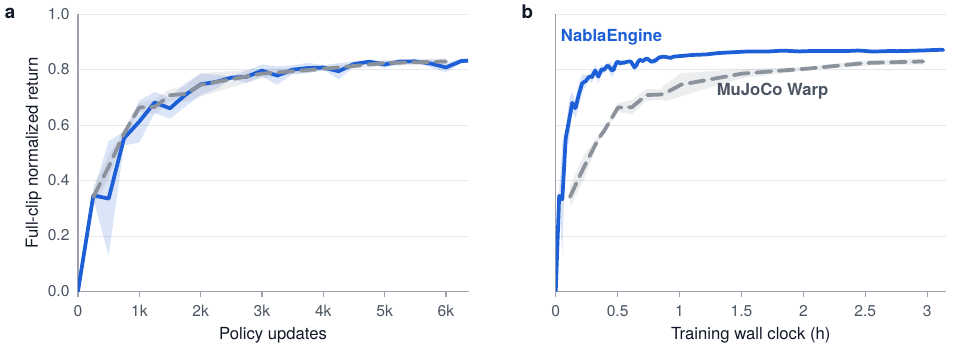}
  \caption{The engine changes speed, not learning. \method (absolute actions) on \engine and on
  the MuJoCo Warp adjoint with the engine's gradient semantics, 3 seeds each (mean $\pm$ sd) on the same GPU, LAFAN1. (a) Per policy update the
  curves overlap. (b) Per hour of training, \engine is faster.}
  \label{fig:operator}
\end{figure}

Is the speed-up the engine's or the algorithm's? Training the absolute-action version of \method on the MuJoCo Warp
adjoint, with \engine{}'s rules for unbounded derivatives ported to it, answers this: per update the two curves overlap
(\cref{fig:operator}a; at matched updates the 3-seed means differ by at most about 0.02, in either direction); per hour
\engine{} is $4.7\times$ ahead (2.02\,h vs 0.43\,h to 4k updates; $4.4\times$ per update,
1.63 vs 0.37\,s). The engine changes speed, not learning.

\subsection{On the physical robot}
\label{sec:real}

\begin{figure}[t]
  \centering
  \includegraphics[width=\textwidth]{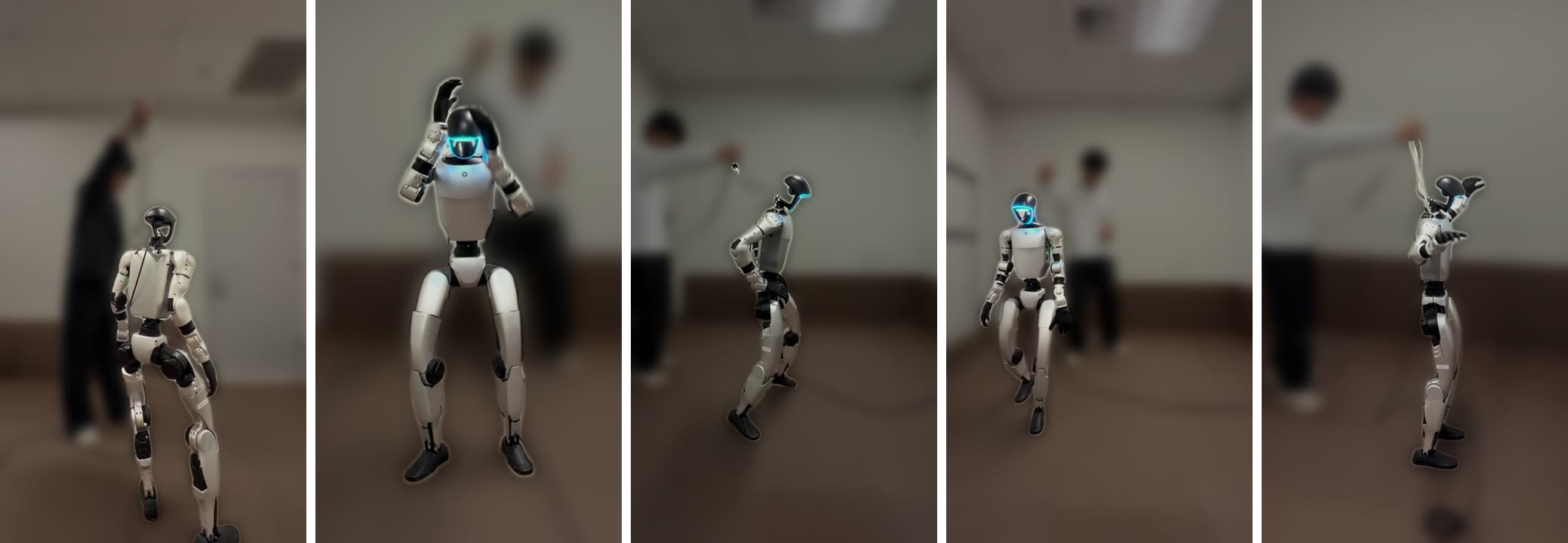}
  \caption{\method policies (recipes earlier than the final one; \cref{sec:real}) on a Unitree G1, one frame from
  each of the five clips (walk, three dances, fighting). Background blurred.}
  \label{fig:real}
\end{figure}

We deployed two checkpoints from recipes earlier than the final one on a Unitree G1: a policy trained for 9.8\,min
on one RTX PRO 6000 (simulation return 0.867) and the averaged weights of seed~2 of the earlier main recipe after
3.0\,h of training (28.5k updates; simulation return 0.911). Both ran unchanged, at the training control rate of
50\,Hz, with no real-world data, system identification or fine-tuning. Each ran the same seven trials over five of
the 40 training clips, a walk, three dances and a fighting sequence of 9 to 35\,s, without a fall (\cref{fig:real}).
The onboard telemetry was logged with the 3.0\,h policy: over the seven trials, the frame-weighted joint position
error is 3.9$^\circ$ RMS. A team member held a slack safety strap.

\section{Limitations and future work}
\label{sec:limitations}

One robot, the Unitree G1, on flat ground is the scope of this version; the kernel design, one world per warp with a
hand-derived adjoint of the whole control step, carries to other robots with their own compile-time specialization, and
uneven terrain needs height fields and further collision types. Our quantitative results are on LAFAN1 in simulation;
on hardware we show two policies of earlier recipes playing five training clips (\cref{sec:real}). We have not yet
trained on larger motion libraries. The gradient holds the
discrete decisions of the contact solve fixed, so it is the exact derivative of the step almost everywhere rather than a
smoothed gradient across contact changes. Per-sample clipping rescales the norms of 9--21\% of the samples per update,
preserving their direction. SuperMimic's official code trains in our physics, which includes self-collisions, while
its stated numbers come from a plant without them; speed comparisons between simulators are made within one GPU type.
Extending \engine{} to more robots and terrains, scaling \method{} to large motion libraries and evaluating the
policies on hardware systematically are the natural next steps.

\section{Conclusion}
\label{sec:conclusion}

First-order policy learning has long promised far more informative updates than zeroth-order methods, but on
contact-rich whole-body control two things held it back: rare unbounded samples in the gradient, and the cost of
computing it. \engine{} removes the second for the Unitree G1: it computes the full MuJoCo dynamics of the robot, all
502 contact pairs included, with analytic gradients that match float64 finite differences, and its adjoint costs less
than another forward step. \method{} removes the first: per-sample median-relative clipping bounds the rare extreme
samples, and multistep regression onto gradient-derived action targets reuses each costly sample with action targets whose
size does not depend on their scale. Together they train a single policy that tracks all 40 LAFAN1 clips and
reaches the 3-hour score of the world-model state of the art in 5.3 minutes on one GPU and levels off at 0.929.

Cutting the cost of a tracking policy from hours to minutes makes training on far larger and more diverse motion
corpora practical, and can thereby accelerate the scaling of humanoid foundation models.

\bibliographystyle{plainnat}
\bibliography{refs}

\appendix

\section{Task details}
\label{app:task}

\paragraph{Data.} The 40 LAFAN1 clips~\citep{harvey2020robust} are the Unitree G1 retarget of the LAFAN1 Retargeting
Dataset~\citep{lafan1retarget2025} (its \texttt{g1/} folder), used without modification. At load time the 30\,FPS clips
are resampled to the 50\,Hz control rate (linear interpolation for root position and joint angles, spherical linear
interpolation for the root orientation), and body poses and velocities are computed by forward kinematics on the
robot model.

\paragraph{Reward.} The reward follows BeyondMimic~\citep{liao2026beyondmimic} with the weights and widths of
SuperMimic~\citep[Table A.2]{liao2026supermimic}. Tracking terms have the form $w \exp(-e/\sigma^2)$, with $e$ the squared
error (orientation: squared rotation angle), averaged over the bodies for the body terms. The body terms are evaluated over 14 bodies, with the reference re-anchored to
the robot's torso in horizontal position and yaw; the anchor terms use the torso itself.
\begin{center}
\small
\begin{tabular}{@{}lcc@{}}
\toprule
Term & $\sigma$ & $w$ \\
\midrule
Body position & 0.3 & 1.0 \\
Body orientation & 0.4 & 1.0 \\
Body linear velocity & 1.0 & 1.0 \\
Body angular velocity & 3.14 & 1.0 \\
Anchor position & 0.3 & 0.5 \\
Anchor orientation & 0.4 & 0.5 \\
Action rate (penalty) & -- & $-0.1$ \\
Joint limits (penalty) & -- & $-2.0$ \\
\bottomrule
\end{tabular}
\end{center}
Every term is multiplied by the control period $\Delta t = 0.02$\,s, so the per-step tracking reward is at most 0.1.
During training only, \method{} adds the termination penalty of \cref{tab:hyper}; evaluation uses the reward above.

\paragraph{Observations.} The actor sees a history of the last 10 proprioception vectors (93 dimensions each): 29 joint
positions relative to the default pose, 29 joint velocities, base angular velocity, gravity direction in the base frame
and the previous action. Uniform noise of SuperMimic's Table A.1 is added during training. The goal consists of the next
10 reference frames, each with 68 features expressed in the reference's own frame (joint positions and velocities,
gravity direction, root angular velocity, root linear velocity in the heading frame, root height), plus the reference
anchor's orientation relative to the robot's anchor (6 dimensions, two rotation-matrix columns), as in BeyondMimic's
variant without state estimation. The critic is privileged: it sees noise-free proprioception, the goal, root errors
and the randomization parameters.

\paragraph{Domain randomization.} Per episode, following SuperMimic's Table A.3: foot friction $\mathcal{U}[0.3,1.2]$,
joint encoder bias $\mathcal{U}[-0.01,0.01]$\,rad and a torso center-of-mass offset. None is applied at evaluation.

\section{Engine details}
\label{app:engine}

\paragraph{Model.} \engine{} simulates the mjlab Unitree G1 model~\citep{zakka2026mjlab}: a free base and 29 actuated
hinge joints, 31 bodies, and 502 contact pairs among capsules, spheres and the floor, self-collision included. The
dynamics follow MuJoCo~\citep{todorov2012mujoco}:
\begin{itemize}
  \item recursive Newton--Euler and composite rigid-body algorithms;
  \item MuJoCo's soft-constraint contact model (solref/solimp) with pyramidal friction cones;
  \item joint limits, dof friction loss and actuator force limits;
  \item Newton's method with exact line search, 10 iterations, tolerance $10^{-8}$;
  \item the implicitfast integrator, 4 substeps of 5\,ms per 20\,ms control step, with the control held;
  \item PD position actuators.
\end{itemize}
Per-world foot friction and torso center-of-mass offsets support domain randomization.

\paragraph{Kernel structure.} One world maps to one thread block of 32 threads (one warp). Within each kernel the world's
state stays in shared memory and registers (6.9\,kB and 8.2\,kB per world for the two forward kernels), and the model's
constants sit in GPU constant memory. Each substep runs two forward kernels:
\begin{itemize}
  \item the first computes kinematics, the bias and mass terms, actuator forces, collision and the constraint rows;
  \item the second runs the Newton solve and the integration.
\end{itemize}
A control step is therefore 8 physics kernels for all worlds; the handoff between the two kernels (mass matrix, forces,
constraint rows and contact Jacobians, about 33\,kB per world) and the state between substeps pass through global
memory. Up to 11--12 worlds are resident per SM on the RTX PRO 6000 and the H100. The whole control step can also run as
one kernel, but it needs the full per-world state (31.9\,kB) in
shared memory, fits fewer worlds per SM, and is slower (forward with tape, 4096 worlds, kernel time: 2.76 vs 1.23\,ms on
the RTX PRO 6000, 4.37 vs 1.76\,ms on the H100, 6.79 vs 3.13\,ms on the RTX 3090, 10.88 vs 5.98\,ms on the V100). The adjoint of a control step is one kernel: one warp per
world reads the tape and runs the derivatives of all four substeps in reverse, with the per-world adjoint state in
shared memory.

\paragraph{Tape and adjoint.} For the gradient, the forward pass also writes a tape of 1,472 float32 values per world and
substep: 23.6\,kB per world and control step, or 96.5\,MB at 4096 worlds. The adjoint replays the contact set,
constraint rows and solver state from it and runs every stage in reverse:
\begin{itemize}
  \item integration;
  \item the constraint solve, differentiated through its optimality conditions with the discrete decisions (contact
    set, active constraints, actuator clamps) held fixed;
  \item collision;
  \item kinematics and dynamics.
\end{itemize}

\paragraph{Constraint-solve adjoint.} Per substep, with state $(q, v)$ and control $u$, let $M(q)$ be the mass matrix,
$c(q,v)$ the bias force, $\tau(q,v,u)$ the actuator force and $a_0 = M^{-1}(\tau - c)$ the unconstrained acceleration.
The constraint solve returns
\begin{equation}
  a^\star = \arg\min_a \tfrac12 (a - a_0)^\top M (a - a_0) + \sum_e s_e(J_e a - r_e),
\end{equation}
over constraint rows $e$ (joint limits and the edges of the pyramidal contact cones), with Jacobians $J_e(q)$, reference
accelerations $r_e(q,v) = -b_e J_e v - k_e \iota_e d_e$, and the penalty $s_e(x) = \tfrac12 D_e x^2$ on the active
side of row $e$ and $0$ otherwise ($D_e$ from the impedance $\iota_e(d_e)$). With the active set $\mathcal{A}$ held
fixed, the optimality condition
\begin{equation}
  F(a, \theta) = M(a - a_0) + \sum_{e\in\mathcal{A}} D_e J_e^\top (J_e a - r_e) = 0, \qquad \theta = (q, v, u),
\end{equation}
is smooth in both arguments, and $\partial F/\partial a = H = M + \sum_{e\in\mathcal{A}} D_e J_e^\top J_e$ is symmetric
positive definite. By the implicit function theorem, $\mathrm{d}a^\star/\mathrm{d}\theta = -H^{-1}\,\partial F/\partial
\theta$, so for an output cotangent $\bar a$
\begin{equation}
  \lambda = H^{-1} \bar a, \qquad \bar\theta = -\nabla_\theta \big[\lambda^\top F(a^\star, \theta)\big]
  \quad (a^\star, \lambda \text{ held constant}).
\end{equation}
Since $M a_0 = \tau - c$,
\begin{equation}
  \lambda^\top F = \lambda^\top \mathrm{ID}(q, v, a^\star) - \lambda^\top \tau(q, v, u)
  + \sum_{e\in\mathcal{A}} D_e (J_e\lambda)(J_e a^\star - r_e),
\end{equation}
where $\mathrm{ID}(q,v,a) = M(q)a + c(q,v)$ is inverse dynamics. $\bar\theta$ is therefore the sum of three gradients of
scalars:
\begin{itemize}
  \item inverse dynamics contracted with $\lambda$, which is one reverse pass of recursive Newton--Euler;
  \item the actuator law contracted with $\lambda$, which is zero for force-clamped actuators in exact mode;
  \item the constraint term, which depends on $q$ through $J_e$, $d_e$ and the contact frame and is differentiated
    through the collision geometry (contact point, normal, distance).
\end{itemize}

The implicitfast integrator ($Q a_{\mathrm{adv}} = M a^\star$ with $Q = M + hK_v$, $v' = v + h a_{\mathrm{adv}}$,
$q' = q \oplus h v'$) is differentiated directly: $y = Q^{-1}\bar a_{\mathrm{adv}}$ and $\bar a^\star = M y$, plus the
$q$-dependence of $M$ through $\nabla_q[y^\top M(q)(a^\star - a_{\mathrm{adv}})]$. A control step chains the 4 substeps
in reverse with $u$ held, summing $\bar u$. Held fixed are:
\begin{itemize}
  \item the contact set and the collision branch;
  \item the active set $\mathcal{A}$, with each pyramid edge separately;
  \item the actuator clamps.
\end{itemize}
The derivative is that of the optimality conditions at the returned iterate $a^\star$, not of the solver iterations; the
warm start receives no gradient.

\paragraph{Gradient modes.} Exact mode (clamp\_kv\,$=1$, normal\,$=0$) is the derivative of the discrete step. The
training default (clamp\_kv\,$=0$, normal\,$=3$) keeps the actuator damping term on force-clamped actuators, freezes
the derivative of a self-contact normal when the two geoms' core distance (between sphere centers or capsule axes) is
below 1\,mm, and holds the segment parameters of near-parallel capsule pairs (below about $2^\circ$) fixed. On held-out states the two modes have the same median gradient errors on
groups G0, G5 and G6; on G4, the states with clamped actuators, the truth has no clamp\_kv\,$=0$ counterpart.

\paragraph{Unbounded derivatives.} Two geometric events of the G1 make the exact derivative of the step unbounded.
\begin{itemize}
  \item \emph{Self-contact at vanishing distance.} The derivative of the contact normal scales as the inverse of the
    distance between the two geoms' cores. One state with nearly touching feet had a step Jacobian norm of about
    $3\times10^8$. On 24 training windows of $H = 64$ (2048 environments each), the gradient grew by more than $10^6$
    through 5 windows, by up to $4.9\times10^{21}$, and 32 rows were not finite. With the normal's derivative frozen
    below 1\,mm, 1 of 24 windows grew by more than $10^6$ (at most $2.2\times10^6$) and no row was non-finite.
  \item \emph{Nearly parallel capsules.} The closest points of two capsule segments are ill-conditioned, roughly as
    $1/\sin^2\theta$. On 172 constructed foot-crossing states, the exact rule amplified a vector--Jacobian product by
    up to $3.6\times10^8$ per control step (median 29); repeated over 64 steps, 40 of the 172 states overflowed to
    non-finite values. With the segment parameters held fixed below about $2^\circ$, the amplification was at most 109
    and no state overflowed; ordinary held-out states reach at most 273.
\end{itemize}
Other events (contact on or off, a constraint row at the edge of its zone, actuator saturation, the solver at its
iteration limit) make the derivative jump rather than diverge. The training default removes the two unbounded terms and, on force-saturated actuators, keeps the actuator damping
term (\cref{app:engine}, gradient modes); on held-out states without these events it equals the exact gradient bit for
bit.

\paragraph{Determinism.} Each world's whole step runs in one warp with a fixed assignment of work to lanes, and worlds
never interact. Reductions are warp shuffles in a fixed order, with no floating-point atomics. The operations the
adjoint replays from the tape use explicitly rounded arithmetic, so the replay equals the forward pass bit for bit.
Outputs and gradients are bitwise identical from run to run: across three processes on 8192 held-out states (V100), and across repeated calls within a process in the verification suite (V100, H100).

\paragraph{Timing protocol.} Times are per 20\,ms control step for 4096 worlds, on held-out states. For \engine{} and
MuJoCo Warp each value is the mean of the medians on two held-out state sets (V100: one set); other simulators use one
held-out state set. Every simulator runs in its own standard configuration, and speed-ups are always within
one GPU. Isaac Lab rows use Isaac Lab's own G1 asset and task defaults; Isaac Sim requires an RT-core GPU, so they are
not available on the H100 and V100. The MuJoCo C reference runs on one AMD EPYC 9555 (16 threads) with MuJoCo's own
thread pool. The MuJoCo Warp gradient times are those of the open adjoint pull request \citep{mjwarp_pr1535} as published; its
public follow-up fixes \citep{mjwarp_adjointfix2026}, which the accuracy tables also report, change the cost by less
than 2\%. The MuJoCo Warp training run of \cref{sec:two-simulators} uses the pull request with these fixes and with
\engine{}'s training gradient rules ported to it.

\begin{figure}[h]
  \centering
  \includegraphics[width=\textwidth]{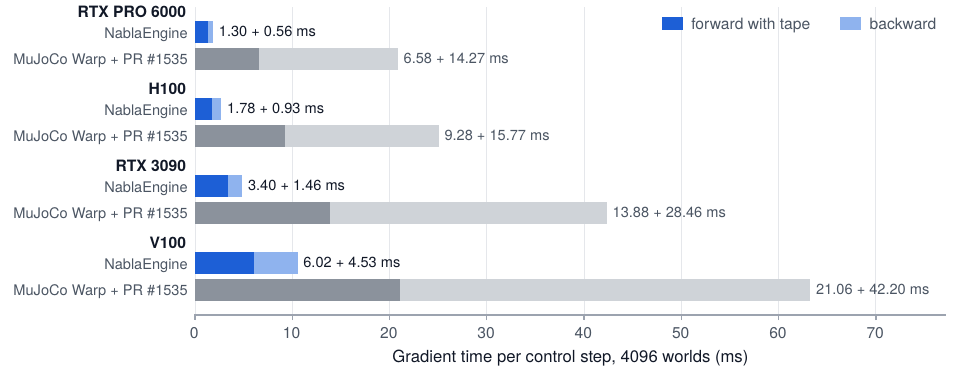}
  \caption{Gradient time per control step for 4096 worlds, split into the forward pass with tape (dark) and the
  backward pass (light), for \engine{} and MuJoCo Warp with the open pull request~\citep{mjwarp_pr1535}. The two parts
  are timed separately, so their sum can differ from the single-call times in \cref{tab:engine-gradient}.}
  \label{fig:engine-split}
\end{figure}

\paragraph{All engines.} \Cref{tab:engine-forward} lists forward times of other GPU engines in their standard
configurations, \cref{tab:engine-gradient} the step-and-gradient times and \cref{fig:engine-split} their split.

\begin{table}[h]
  \centering
  \footnotesize
  \caption{Forward time per control step, 4096 worlds, ms (\engine{} and MuJoCo Warp: mean of the medians on two
  held-out state sets, V100 one set; other simulators: one held-out state set). n/s: not supported, because Isaac Sim
  requires an RT-core GPU; ---: not measured. Speed-ups hold within one GPU.}
  \label{tab:engine-forward}
  \setlength{\tabcolsep}{4pt}
  \begin{tabular}{lrrrrrr}
    \toprule
    Engine & RTX 5090 & RTX PRO 6000 & B300 & H100 & RTX 3090 & V100 \\
    \midrule
    \engine & \textbf{1.12} & \textbf{1.23} & \textbf{1.47} & \textbf{1.72} & \textbf{3.42} & \textbf{5.79} \\
    MuJoCo Warp 3.17.0 & 5.46 & 6.17 & 7.00 & 8.69 & 12.88 & 18.4 \\
    Genesis 1.4.5 & --- & 13.45 & --- & 16.36 & 36.73 & 139.11 \\
    MJX 3.17 (same model) & --- & 255.0 & --- & 216.0 & 510.3 & 710.4 \\
    MJX 3.17 (Menagerie G1, fewer contacts) & --- & 50.7 & --- & 58.4 & 129.2 & 150.7 \\
    Isaac Lab 3.0, Newton (3 box colliders) & --- & 12.9 & --- & n/s & 35.2 & n/s \\
    Isaac Lab 3.0, PhysX (3 box colliders) & --- & 27.5 & --- & n/s & 45.2 & n/s \\
    Isaac Lab 3.0, PhysX (38 shapes, self-collision) & --- & 54.1 & --- & n/s & 87.5 & n/s \\
    MuJoCo C 3.17 (CPU, float64; AMD EPYC 9555, 16 threads) & \multicolumn{6}{c}{34.7 (CPU)} \\
    \bottomrule
  \end{tabular}
\end{table}

\begin{table}[h]
  \centering
  \small
  \caption{One control step and its gradient from the input state, 4096 worlds, ms (same timing basis as
  \cref{tab:engine-forward}).}
  \label{tab:engine-gradient}
  \begin{tabular}{lrrrrrr}
    \toprule
    & RTX 5090 & RTX PRO 6000 & B300 & H100 & RTX 3090 & V100 \\
    \midrule
    \engine (forward with tape + adjoint) & \textbf{1.64} & \textbf{1.78} & \textbf{2.34} & \textbf{2.74} & \textbf{5.26} & \textbf{10.6} \\
    MuJoCo Warp + PR \#1535 (adjoint graph) & 18.96 & 20.98 & 19.25 & 25.13 & 42.74 & 63.2 \\
    Speed-up & $11.6\times$ & $11.8\times$ & $8.2\times$ & $9.2\times$ & $8.1\times$ & $6.0\times$ \\
    \bottomrule
  \end{tabular}
\end{table}

\section{Full accuracy tables}
\label{app:accuracy}

Held-out benchmark: four corpora of 4096 states each (16,384 in total). The truth is MuJoCo 3.17 C in float64, with the
same model and per-world randomization, the Newton solver at tolerance $10^{-13}$ and up to 2000 iterations, and one
control step from each state. Gradient errors are relative errors of vector--Jacobian products with three stored random
output directions per state (i.i.d.\ standard-normal entries, fixed seed, unnormalized), against float64 central finite
differences at step $10^{-6}$, floored per state at the truth's own precision (the difference between the finite
differences at steps $10^{-7}$ and $10^{-6}$, typically about $10^{-7}$).

\begin{table}[h]
  \centering
  \small
  \caption{Accuracy on held-out states against MuJoCo C 3.17 in float64. Forward: median relative qvel error.
  Gradient: median relative VJP error against float64 central finite differences, for position / velocity / action.
  The truth itself is known to about $10^{-7}$.}
  \label{tab:engine-accuracy}
  \begin{tabular}{lccc}
    \toprule
    & \multicolumn{3}{c}{Gradient (median relative VJP error)} \\
    \cmidrule(lr){2-4}
    Engine & Position & Velocity & Action \\
    \midrule
    \engine & $3.6\times10^{-5}$ & $1.0\times10^{-5}$ & $1.9\times10^{-6}$ \\
    MJX, forward-mode Jacobian & $9.6\times10^{-5}$ & $2.1\times10^{-5}$ & $2.9\times10^{-6}$ \\
    MuJoCo C finite differences (step $10^{-6}$) & $6.7\times10^{-4}$ & $1.5\times10^{-4}$ & $7.8\times10^{-6}$ \\
    Genesis (velocity only) & --- & $2.3\times10^{-1}$ & --- \\
    MuJoCo Warp, open pull request & 0.29 & 0.34 & 0.015 \\
    \begin{tabular}[t]{@{}l@{}}MuJoCo Warp, with the follow-up fix\\(etaoxing/mujoco\_warp\#1)\textsuperscript{a}\end{tabular}
      & $5.3\times10^{-5}$ & $1.0\times10^{-5}$ & $2.1\times10^{-6}$ \\
    \midrule
    & \multicolumn{3}{c}{Forward (median relative qvel error)} \\
    \midrule
    \engine & \multicolumn{3}{c}{$8.5\times10^{-7}$} \\
    MuJoCo Warp & \multicolumn{3}{c}{$1.3\times10^{-6}$} \\
    MJX & \multicolumn{3}{c}{$2.9\times10^{-6}$} \\
    Genesis & \multicolumn{3}{c}{$5.5\times10^{-2}$} \\
    \bottomrule
  \end{tabular}
  \par\smallskip\footnotesize\textsuperscript{a}\,\citet{mjwarp_adjointfix2026}.
\end{table}

\paragraph{Groups.} Group labels come from MuJoCo C alone, per control step (any substep). A state is \emph{smooth} if
none of the following occurs: a self-contact core distance below 5\,mm; a contact flip (the contact set changes, or a
distance below $10^{-6}$\,m); a kink (a constraint row at the edge of its zone,
$|x_e| / (|J_e a| + |r_e|) < 10^{-4}$); a contact frame near its basis switch; the solver at its iteration limit.
\begin{itemize}
  \item G0: smooth, no clamped actuator.
  \item G4: smooth, a clamped actuator (PD force at or beyond its range).
  \item G5: smooth, an active joint-limit row, no clamped actuator.
  \item G6: smooth, a contact between two robot geoms (self contact), no clamped actuator.
\end{itemize}
Accuracy is verified on every state at which a derivative exists: smooth states with and without self-contact, active
joint limits and clamped actuators. At non-smooth states the derivative of the step jumps or is undefined and finite
differences measure nothing there; what matters at those states is that no output is non-finite, which holds on all
16,384 states. G5 and G6 are subsets of G0. The forward errors use every state of a group on which the model is valid. Gradient errors
use up to 64 states per group and corpus (seed 0), 256 per group pooled. \Cref{sec:engine} reports the forward error over
all states and the gradient error on G0. With either gradient mode, \engine{} produced no non-finite output or gradient
on any of the 16,384 states. Both modes share one forward pass, which is bitwise identical.

\begin{table}[h]
\centering
\small
\caption{Forward error per group, 10 solver iterations: median / 99th percentile of the relative qvel error, and in
parentheses the 99th percentile of the qpos error (rad). States per group: G0 13,480, G4 476, G5 5,645, G6 5,668. MuJoCo
C in float64 with 10 iterations has median 0 and 99th percentile at most $1.9\times10^{-15}$ in every group.}
\label{tab:acc-forward-groups}
\setlength{\tabcolsep}{4pt}
\resizebox{\linewidth}{!}{%
\begin{tabular}{@{}lllll@{}}
\toprule
 & G0 & G4 & G5 & G6 \\
\midrule
\engine{} & 8.6e-7 / 4.7e-6 (5.4e-7) & 4.6e-7 / 6.6e-6 (1.8e-6) & 8.4e-7 / 5.5e-6 (6.5e-7) & 9.4e-7 / 6.9e-6 (1.0e-6) \\
MuJoCo Warp & 1.3e-6 / 8.8e-6 (1.1e-6) & 6.3e-7 / 7.2e-6 (1.5e-6) & 1.2e-6 / 6.8e-6 (9.2e-7) & 1.3e-6 / 8.8e-6 (9.9e-7) \\
MuJoCo Warp, open PR & 1.3e-6 / 9.3e-6 (1.2e-6) & 6.4e-7 / 1.0e-5 (3.4e-6) & 1.2e-6 / 8.6e-6 (1.2e-6) & 1.4e-6 / 1.1e-5 (1.4e-6) \\
MJX (float32) & 2.7e-6 / 6.3e-4 (7.2e-5) & 1.3e-6 / 4.2e-4 (1.1e-4) & 2.6e-6 / 5.1e-4 (6.1e-5) & 5.5e-6 / 6.6e-4 (7.3e-5) \\
Genesis (float32) & 4.9e-2 / 4.8e-1 (6.0e-2) & 5.7e-2 / 5.4e-1 (1.4e-1) & 4.9e-2 / 4.8e-1 (6.1e-2) & 1.0e-1 / 5.1e-1 (7.1e-2) \\
\midrule
\multicolumn{5}{@{}l}{\emph{Reference: MuJoCo Warp against itself, run to run}} \\
 & 3.1e-7 / 3.2e-6 (4.3e-7) & 3.0e-7 / 2.2e-6 (8.0e-7) & 3.4e-7 / 3.3e-6 (3.8e-7) & 3.4e-7 / 3.5e-6 (3.9e-7) \\
\bottomrule
\end{tabular}}
\end{table}

\begin{table}[h]
\centering
\small
\caption{Gradient error per group and input block, 10 solver iterations, 256 states per group: median floored relative
VJP error for position / velocity / action. Exact and default denote \engine{}'s gradient modes. ``n/c'': not
comparable (the truth is the exact derivative and has no counterpart of the default mode on clamped actuators);
``---'': not supported.}
\label{tab:acc-gradient-groups}
\setlength{\tabcolsep}{4pt}
\resizebox{\linewidth}{!}{%
\begin{tabular}{@{}lllll@{}}
\toprule
 & G0 & G4 & G5 & G6 \\
\midrule
\engine{}, exact & 3.6e-5 / 1.0e-5 / 1.9e-6 & 4.3e-5 / 1.3e-5 / 2.0e-6 & 3.8e-5 / 1.1e-5 / 1.8e-6 & 3.1e-5 / 1.0e-5 / 2.3e-6 \\
\engine{}, default & same as exact & n/c & same as exact & same as exact \\
MJX, forward-mode Jacobian & 9.6e-5 / 2.1e-5 / 2.9e-6 & 1.2e-4 / 2.3e-5 / 3.0e-6 & 9.4e-5 / 2.0e-5 / 3.1e-6 & 1.0e-4 / 2.5e-5 / 4.8e-6 \\
MuJoCo C, finite differences & 6.7e-4 / 1.5e-4 / 7.8e-6 & 7.1e-4 / 1.8e-4 / 7.2e-6 & 6.6e-4 / 1.5e-4 / 7.9e-6 & 5.5e-4 / 1.3e-4 / 7.8e-6 \\
Genesis (velocity only) & --- / 2.3e-1 / --- & --- / 3.4e-1 / --- & --- / 2.7e-1 / --- & --- / 3.4e-1 / --- \\
MuJoCo Warp, open PR & 2.9e-1 / 3.4e-1 / 1.5e-2 & 5.0e-1 / 4.8e-1 / 8.9e-2 & 3.1e-1 / 3.5e-1 / 1.8e-2 & 3.5e-1 / 4.0e-1 / 1.9e-2 \\
MuJoCo Warp, follow-up fix & 5.3e-5 / 1.0e-5 / 2.1e-6 & n/c & 5.1e-5 / 1.1e-5 / 1.9e-6 & 4.5e-5 / 1.0e-5 / 2.3e-6 \\
\bottomrule
\end{tabular}}
\end{table}

MuJoCo C finite differences are forward differences with step $10^{-6}$, chained over the four substeps.
Versions: MuJoCo Warp~\citep{mujocowarp2025} and MJX~\citep{mjx2024} 3.17.0, Genesis~\citep{genesis2024} 1.4.5. The
open pull request is \citet{mjwarp_pr1535}, the follow-up fix \citet{mjwarp_adjointfix2026}.

\section{Hyperparameters and baselines}
\label{app:hyper}

\begin{table}[h]
\centering
\small
\caption{\method{} hyperparameters.}
\label{tab:hyper}
\begin{tabular}{@{}lp{0.66\linewidth}@{}}
\toprule
Item & Setting \\
\midrule
Environments & 2048 BPTT training windows per update \\
Window length & $H = 64$ simulated steps (1.28\,s) \\
Discount, TD & $\gamma = 0.99$, $\lambda = 0.95$ \\
Actor & encoder--decoder MLP 2048--1024--512--256; anchor orientation into the decoder; LayerNorm and SiLU in every
hidden layer; \textmu P-style initialization \\
Critic & MLP 256--128 with LayerNorm and SiLU, fit on the pre-transition state \\
Optimizers & actor hidden matrices: Muon (RMS step 0.1); all other parameters: Adam \\
Actions & residual: offset to the reference joint pose \\
Learning rate & $2\times10^{-4}$, 100-update warmup, $3\times$ for the actor over updates 0--500 (back to $1\times$
by 800), constant to update 1{,}000, cosine decay to $0.1\times$ at 16{,}000, policy steps capped at $0.02\times$ from
16{,}000 \\
Action targets & RMS setpoint change 0.03\,rad (decays to $0.3\times$ with the LR); 8 regression steps \\
 & of 16{,}384 samples; separate Adam, lr $6\times10^{-4}$ \\
Per-sample clipping & $\kappa = 3\times$ batch median of the action-cotangent norm \\
Termination penalty & $W = 2$, $dz_0 = 0.08$, $d\theta_0 = 0.10$ \\
Episode ends & reset and continue the window; a fall bootstraps nothing (no survival mask) \\
Policy averaging & EMA of the actor, decay $\min(0.999, (1+n)/(10+n))$; the averaged policy is evaluated \\
Window action noise & 0.1 \\
Start states & replay of $1000\times4096$ states from a data collector (action noise 0.1, update-to-data 0.4) \\
Reset sampling & return-based bins; from update 300, 15\% to the 10\,s before falls found by a deterministic probe
of the averaged policy every 150 updates \\
Physics & \engine{}, 10 solver iterations, FP32; network matrix products bfloat16 \\
Updates & 24{,}000 \\
Time per update & 0.25\,s on one RTX PRO 6000 \\
\bottomrule
\end{tabular}
\end{table}

\paragraph{Sensitivity.} \Cref{tab:sensitivity} varies one setting of the residual recipe at a time. In the earlier main recipe, a 500-update learning-rate warmup instead of 100 trains more slowly (0.82 after 9.3 instead of 7.9\,min) but converges slightly better (0.913 against 0.910 at 2.99\,h, 3 seeds each).

\begin{table}[h]
\centering
\small
\caption{Sensitivity of the earlier residual recipe (masked, no policy averaging): full-clip normalized return after 1.5\,h of training on one RTX PRO 6000,
mean $\pm$ sd over 3 seeds (4 for $H = 32$). Recipe values in bold. Diverged: training stopped by the divergence guard.}
\label{tab:sensitivity}
\begin{tabular}{@{}llll@{}}
\toprule
Setting & Value & Return after 1.5\,h & Note \\
\midrule
Clipping threshold $\kappa$ & 1 & 0.853 $\pm$ 0.008 & \\
 & \textbf{3} & \textbf{0.869 $\pm$ 0.003} & \\
 & 10 & 0.849 $\pm$ 0.011 & \\
 & 30 & 0.849 $\pm$ 0.005 & \\
 & none & --- & 3 of 3 diverged \\
Gradient rules & \textbf{\engine{} rules} & \textbf{0.869 $\pm$ 0.003} & \\
 & exact everywhere & 0.867 $\pm$ 0.002 & \\
Window length $H$ & 16 & --- & 3 of 3 diverged (critic value runaway) \\
 & 32 & 0.815--0.841 & 1 of 4 collapsed (critic value runaway) \\
 & \textbf{64} & \textbf{0.869 $\pm$ 0.003} & about 14.3k updates \\
 & 128 & 0.839 $\pm$ 0.004 & about 7.9k updates \\
Windows per update & 1024 & 0.865 $\pm$ 0.002 & about 27.0k updates \\
 & 2048 & 0.867 $\pm$ 0.004 & about 20.3k updates \\
 & \textbf{4096} & \textbf{0.869 $\pm$ 0.003} & about 14.3k updates \\
 & 8192 & 0.852 $\pm$ 0.002 & about 8.6k updates \\
\bottomrule
\end{tabular}
\end{table}

\paragraph{Ablations of the final recipe.} \Cref{tab:abl-final} changes one component of the final recipe at a
time. The short block runs every arm for about 54 minutes on one A40 (3 seeds, evaluated every 200 updates); the long
block trains the two arms closest to the control for the full 24k updates on one RTX PRO 6000 (3 seeds). Without
per-sample clipping the regression optimizer keeps its global-norm clip, and every seed collapses within about 100
updates: the scale $\delta / \operatorname{RMS}(g)$ falls from about $10^4$ to between 0.6 and 300 as a few samples with
very large gradients come to dominate the RMS. A single regression step and a fixed gain in place of the RMS
normalization both train stably and reach the levels later. Clipping only the window's action gradients and holding
$\delta$ fixed match the control within seed noise over the full run.

\begin{table}[h]
\centering
\small
\caption{Ablations of the final recipe on LAFAN1, one change at a time, 3 seeds each, scored on MuJoCo Warp. Hold: the
first evaluation after which every later evaluation stays at or above the level (minutes; in parentheses the number of
seeds that hold it if fewer than 3). Fixed gain: a constant gain equal to the control's early median of
$\delta/\operatorname{RMS}(g)$, with the same $\delta$ schedule. Fixed $\delta$: no decay of the step size. In the short
block $\delta$ has decayed only to $0.87\,\delta_0$ by the last checkpoint, so it cannot show the effect of the decay.}
\label{tab:abl-final}
\begin{tabular}{@{}lcccc@{}}
\toprule
\multicolumn{5}{@{}l}{\emph{About 54 minutes, one A40}} \\
Arm & Hold 0.866 & Hold 0.890 & Return, 30\,min & Return, last \\
\midrule
Final recipe & 13.9\,$\pm$\,1.3 & 38.7\,$\pm$\,3.3 & 0.887 & 0.897 \\
No per-sample clipping & never & never & 0.034 & 0.058 \\
Single regression step ($M = 1$) & 21.4\,$\pm$\,0.8 & 51.7\,$\pm$\,2.1 (2) & 0.877 & 0.889 \\
Fixed gain instead of RMS normalization & 20.0\,$\pm$\,1.3 & 43.4 (1) & 0.878 & 0.888 \\
Clipping of the window's action gradients only & 13.1 & 39.3 & 0.884 & 0.900 \\
Fixed $\delta$ & 12.5 & 40.9 & 0.884 & 0.899 \\
\midrule
\multicolumn{5}{@{}l}{\emph{24k updates, one RTX PRO 6000}} \\
Arm & Hold 0.866 & Hold 0.890 & Hold 0.929 & Return, updates 21k--24k \\
\midrule
Final recipe & 5.6 & 17.4 & 83.5 (2) & 0.9315\,$\pm$\,0.0015 \\
Clipping of the window's action gradients only & 6.4 & 16.1 & 88.1 & 0.9297\,$\pm$\,0.0025 \\
Fixed $\delta$ & 6.0 & 16.7 & 77.7 & 0.9315\,$\pm$\,0.0008 \\
\bottomrule
\end{tabular}
\end{table}

\paragraph{Network architecture.} The actor follows SuperMimic's encoder--decoder design \citep{liao2026supermimic}.
A motion encoder maps the next 10 reference frames (68 features each) through an MLP with hidden widths
2048--1024--512--256 to a 32-dimensional latent code with a tanh output. A decoder with the same hidden widths maps the
latent code, the 10-frame proprioception history (93 features per frame) and the 6-dimensional anchor orientation to
the 29 joint offsets. The critic is an MLP with hidden widths 256--128 on privileged inputs (\cref{app:task}).

The design encodes the reference motion into a compact latent space and lets the decoder act as a general motor
policy conditioned on that code. This is the structure along which tracking policies are scaled toward foundation models
for humanoid control: the latent code is the interface through which motion generators, teleoperation or downstream
policies command the tracker, and the same decoder serves all motions
\citep{chen2025gmt,liao2026beyondmimic,luo2026sonic}. We keep it, rather than a monolithic MLP, so that a policy
trained with \method{} can be used the same way.

\paragraph{Baselines.}
\begin{itemize}
  \item \textbf{PPO, BeyondMimic configuration}~\citep{liao2026beyondmimic,schulman2017proximal}: on \engine{} (forward only), 4096
    environments $\times$ 24 steps per iteration, 5 epochs $\times$ 4 minibatches, lr $10^{-3}$ (adaptive, KL 0.01),
    clip 0.2, entropy 0.005, $\gamma = 0.99$, $\lambda = 0.95$, MLP 512--256--128, BeyondMimic's adaptive sampling;
    6\,h budget. One policy is trained on all 40 clips; BeyondMimic itself trains one policy per motion, so this row
    uses its configuration and is not a reproduction of its results.
  \item \textbf{PPO, SuperMimic actor and reset sampling}: as above with 8192 environments, SuperMimic's encoder--decoder
    actor (2048--1024--512--256) and critic (256--128), and return-based reset sampling; 6\,h budget.
  \item \textbf{SuperMimic, official code}~\citep{liao2026supermimic}: see \cref{app:supermimic}.
  \item \textbf{SHAC}~\citep{xu2022shac}: see \cref{app:shac}.
\end{itemize}
All baselines use the same environment, observation and evaluation reward as \method{}. The smooth termination penalty
(\cref{eq:term-penalty}) is part of our recipe and does not account for the gaps: removing all of fall handling, the
penalty included, changes the return of \method{} by at most 0.01 at matched wall clock (3 seeds, both recipes).

\section{SuperMimic reproduction and cost estimate}
\label{app:repro}

\paragraph{Reproduction.} The gradient analysis of \cref{sec:gradients} uses a public reproduction of SuperMimic
(\url{https://github.com/CuboiLeo/SuperMimic-Reproduced}) with the hyperparameters stated in its paper: windows of 32
imagined steps for 4096 rollouts per update, an update-to-data ratio of 0.4, and a world model that is an MLP with three
hidden layers of 2048 units and SiLU activations, predicting the change of the canonicalized state from the state, the
action and the simulation context. Its policy observes the anchor orientation of the reference as ours does. We trained
one run on an RTX PRO 6000 and evaluated its policy with our protocol (\cref{app:eval}); it reaches a full-clip return
of 0.789 after 2.88\,h. It follows the hyperparameters SuperMimic states and serves one purpose: to provide world models and policies at
several stages of training on which exact and surrogate gradients can be compared along the same trajectories. It is
not a baseline, and its checkpoints at 0.44,
1.14, 1.83 and 2.88\,h provide the policies and world models of \cref{sec:gradients}.

\paragraph{Cost estimate.} SuperMimic states 320,000 training iterations and that the world model and the policy are
updated in every iteration. Its BONES-SEED training curve, with a checkpoint every 4,000 iterations, ends at 9.37\,h on one
H100, which gives about 34,000 iterations per hour, or about 0.1\,s per iteration; we assume that LAFAN1 iterations cost
the same. Reaching 0.82 within 1.2\,h then corresponds to about 41,000 iterations. If one instead counts policy updates at
the stated update-to-data ratio of 0.4, as the public reproduction does, these become about 16,000 updates of about
0.26\,s each, and \method{}'s 1,250 updates to 0.82 are about one thirteenth of them rather than one thirty-third.

\section{SuperMimic official code}
\label{app:supermimic}

We train SuperMimic from its official release (commit abc124c) in our physics: the mjlab G1 with all collision pairs,
self-collisions included, 10 solver iterations, and all 40 LAFAN1 clips, with proprioceptive actor input and a 10-frame
observation history as in its paper, and every other setting as released. Three seeds train for about 3.2\,h each,
seed~1 on an RTX PRO 6000 and seeds 0 and 2 on H100; times are converted to the RTX PRO 6000 by update count at
0.315\,s per update. Its policies are scored with our protocol on MuJoCo Warp (\cref{app:eval}). \Cref{tab:supermimic}
gives the per-seed results. All three seeds pass 0.82 within 47 minutes and then rise slowly; seed~2 reached 0.847
after 168 minutes and fails 2 to 4 clips in three evaluations of its last checkpoint, which lowers its plateau. The
per-seed times in \cref{tab:supermimic} are the first evaluated checkpoint at or above each level; the seed-mean curve
used for \cref{tab:levels} crosses 0.721 after 16.1 and 0.82 after 45.5\,min.

\begin{table}[h]
\centering\small
\caption{SuperMimic's official code in our physics, per seed: minutes on the RTX PRO 6000 clock (converted by update
count) to the first evaluation at or above 0.721 and 0.82, plateau (mean over the last 30 minutes) and best checkpoint,
scored on MuJoCo Warp. Last row: first crossing of the seed-mean curve, as in \cref{tab:levels}.}
\label{tab:supermimic}
\begin{tabular}{@{}lcccc@{}}
\toprule
Seed (GPU) & To 0.721 (min) & To 0.82 (min) & Plateau & Best \\
\midrule
0 (H100) & 18.4 & 47.2 & 0.851 & 0.852 \\
1 (RTX PRO 6000) & 13.1 & 42.0 & 0.850 & 0.851 \\
2 (H100) & 15.7 & 42.0 & 0.815 & 0.848 \\
\midrule
Mean $\pm$ sd & 15.7\,$\pm$\,2.7 & 43.7\,$\pm$\,3.0 & 0.838\,$\pm$\,0.021 & 0.850\,$\pm$\,0.002 \\
Seed-mean curve & 16.1 & 45.5 & --- & --- \\
\bottomrule
\end{tabular}
\end{table}

\section{Cross-simulator evaluation}
\label{app:sim2sim}

\Cref{tab:sim2sim} scores the same checkpoints on both simulators, with the same robot model (self-collision on), the
same 10 solver iterations and the protocol of \cref{sec:setup}. Policies that are still early in training, which drop
an occasional clip, score up to about 0.01 higher in the simulator they were trained in; at 3\,h the \engine{}-trained
policies transfer to MuJoCo Warp with no measurable loss, and so do the final \method policies after 24k updates
(per-seed differences at most 0.001). For scale, enabling the robot's self-collisions alone moves an
InstantTrack policy from 0.87 to 0.75 (\cref{app:instanttrack}).

\begin{table}[h]
\centering\small
\caption{The same checkpoints scored on MuJoCo Warp and on \engine{} (mean $\pm$ sd over 3 seeds, at the last
evaluated checkpoint at or before the given update count). Re-scoring on MuJoCo Warp alone varies by up to about 0.004
without falls and by 0.01--0.04 when a clip falls.}
\label{tab:sim2sim}
\begin{tabular}{llrccc}
\toprule
Policy & Trained on & Updates & On MuJoCo Warp & On \engine{} & Difference \\
\midrule
\method, absolute actions & \engine{} & 6{,}000 & 0.809\,$\pm$\,0.017 & 0.819\,$\pm$\,0.018 & $+$0.010 \\
\method, absolute actions & MuJoCo Warp & 6{,}000 (3\,h) & 0.830\,$\pm$\,0.011 & 0.817\,$\pm$\,0.005 & $-$0.014 \\
\method, absolute actions & \engine{} & 29{,}000 (3\,h) & 0.872\,$\pm$\,0.005 & 0.872\,$\pm$\,0.005 & 0.000 \\
\method, earlier recipe & \engine{} & 6{,}000 & 0.878\,$\pm$\,0.012 & 0.878\,$\pm$\,0.012 & 0.000 \\
\method, earlier recipe & \engine{} & 29{,}000 (3\,h) & 0.917\,$\pm$\,0.002 & 0.917\,$\pm$\,0.002 & 0.000 \\
\method, final recipe (5 seeds) & \engine{} & 24{,}000 (1.7\,h) & 0.930\,$\pm$\,0.004 & 0.930\,$\pm$\,0.003 & 0.000 \\
\bottomrule
\end{tabular}
\end{table}

\section{InstantTrack details}
\label{app:instanttrack}

We run InstantTrack \citep{dao2026instanttrack} in version 1.4.0, its latest release (commit b9512a9 of 7 November
2026), with its released training code and settings. Only the plant differs: it trains with \engine{} as its
simulator on the shared G1 model, which is the mjlab G1 model with capsule collision geometry, self-collisions on and 10 solver iterations. Each of
its 3 seeds trains for 3\,h on the same RTX PRO 6000 as \method, with 32 CPU cores, at 15.8 iterations per second, and
its averaged (EMA) policy is scored with our protocol (\cref{app:eval}). In its original plant, without
self-collisions and scored by its own evaluator, the same code reaches $0.900\pm0.001$ over the last 30 minutes of 3\,h
(3 seeds), close to the curve its author reports.

InstantTrack policies trained in their original plant score 0.87 there and 0.75 in the \engine{} plant. Enabling the same robot's self-collisions in MuJoCo Warp alone gives 0.75, while the collision
geometry accounts for about 0.02 and the solver iteration count (10 against 100) has no measurable effect.

\section{World-model gradients with \method's components}
\label{app:wm-grad}

\paragraph{Noise-free imagined windows.} This section uses InstantTrack version 1.1. InstantTrack collects its data with action noise $\sigma=0.1$ but imagines
its training windows without noise. In that setting the policies it trains in our physics develop a failure that shows
only in deterministic execution: at 4\,h its own policy scores 0.804 with 4 of 40 clips failed when executed
deterministically, and 0.866 with $\sigma=0.1$ executed noise (one seed). With \method's components on its world-model
gradients the same failure appears earlier: after about 0.6\,h with clipping and action targets together (0.094,
39 clips failed, against 0.821 with noise), after 2--3\,h with clipping alone and after about 3.5\,h with action targets
alone. The public SuperMimic reproduction with both components shows the same signature (0.643 with all 40 clips failed,
against 0.810 with noise). Imagined windows with action noise remove it: with $\sigma=0.05$ the failure is delayed, and
with $\sigma=0.1$ no seed shows it; these are the version-1.1 runs of \cref{sec:main-result}. We did not analyze the mechanism.
\method's windows always carry action noise ($\sigma=0.1$ in the final recipe, 0.05 in the earlier ones), and its policies behave as expected: deterministic
execution gives the highest return in every checkpoint we tested (3\,h, and 4, 5 and 6\,h of longer runs; for example
0.907--0.911 deterministic against 0.871--0.878 with $\sigma=0.1$ at 3\,h, 3 seeds).

\paragraph{A weaker world model.} In our own code, with a world model trained alongside the policy on the same noisy
windows, the world-model gradient agrees much less with the exact one: median cosine 0.08 early in training and
0.40--0.47 later, with 13--30\% of the samples pointing the opposite way. With \method's components, training on it
reaches 0.80 and 0.86 after 3\,h (two seeds), against 0.91 on exact gradients (earlier recipe); without the components, 0.80 on
world-model and 0.81 on exact gradients (two seeds each). The size of the per-step advantage of exact gradients thus depends on the quality of the surrogate: small
against a well-trained world model (\cref{sec:main-result}), large against a weak one.

\section{SHAC details}
\label{app:shac}

Both SHAC variants use the same environment, actor, critic, 4096 environments, learning-rate schedule and window noise as
\method{}. SHAC's algorithm:
\begin{itemize}
  \item $H = 32$ windows continued across updates, with no replay starts and no collector;
  \item the actor's own window gradient, one Adam step per window;
  \item no per-sample clipping and no termination penalty;
  \item a TD($\lambda = 0.95$) critic, empirical observation normalization and uniform reset sampling.
\end{itemize}
\textbf{SHAC (published guard)} clips the actor gradient to global norm 1.0 and stops training at the first update whose
norm before clipping is not finite or exceeds $10^6$, as in the published code. \textbf{SHAC (changed guards)} instead
clips to global norm 1.0 and then element-wise to 0.1, as SuperMimic does, and skips non-finite updates instead of
stopping. Both run on \engine{}; the
published-guard variant also runs on MuJoCo Warp with the engine's gradient semantics. They differ from \method{} in
several components at once, so they are baselines, not single-component ablations.

\section{Evaluation protocol}
\label{app:eval}

Each of the 40 LAFAN1 clips is played once from its first reference frame with 10 solver iterations, a deterministic
actor (the mean action), no domain randomization and no observation or action noise. Every run is played on MuJoCo
Warp; \cref{app:sim2sim} scores the same checkpoints on \engine{}. The
clip's return is the sum of the per-step reward over its steps until the clip ends or the first termination (height
error above 0.25\,m or tilt above 0.8). The per-step reward is the six tracking terms minus the action-rate and
joint-limit penalties. The sum is divided by the clip's maximum attainable tracking reward, and the score is the
unweighted mean over the 40 clips. A clip with a fall therefore keeps the reward earned before the fall and scores zero
afterwards. This matches the normalized return defined by \citet{liao2026supermimic}. Failed clips are clips with a
fall.

The body linear-velocity term takes the reference and the robot velocity of each body at the same point, the body
origin, as BeyondMimic's mjlab implementation does, so a perfect tracker scores 1. A center-of-mass variant, which
compares the reference velocity at each body's center of mass with the robot's velocity at the body origin, caps a
perfect tracker at 0.991; scoring the same checkpoints on the body-origin measure gives returns about 0.003 to 0.007
higher, depending on the method. The final recipe trains with the center-of-mass variant of this reward term; its scores, like all others, use the
body-origin measure. All scores use the body-origin measure, except the following studies of earlier
recipes, which use the center-of-mass measure: the sensitivity study (\cref{tab:sensitivity}), the effect of policy averaging and unmasked
windows (\cref{sec:method}), the update-count estimate of
\cref{sec:gradients}, and the analyses with InstantTrack version 1.1 and with our own world model
(\cref{sec:main-result,app:wm-grad}).

Every checkpoint of the first training hour is evaluated (every 250 updates), then every 1000 updates, plus the last.
Training wall clock excludes start-up compilation and all evaluation. The plateau of a run is, per seed, the mean
return over the checkpoints evaluated in the last 30 minutes of its training budget, then the mean $\pm$ sd over seeds.
Times to a score are first crossings of the seed-mean curve, linearly interpolated between evaluations and counted from
the first training update (for InstantTrack, whose clock starts before its first compilation, that compilation is
subtracted).

\section{Tracking metrics and drift}
\label{app:tracking}

\Cref{tab:tracking-metrics,tab:drift} report the final \method policies against the two world-model trackers in the
same physics, on MuJoCo Warp with 10 solver iterations.

\begin{table}[h]
\centering
\small
\caption{Tracking quality of the final policies on LAFAN1 (all 40 clips, each played once from its first frame,
deterministic, MuJoCo Warp). Return: full-clip normalized return; success: share of clips completed without a fall;
MPJPE: mean position error of the 14 tracked bodies relative to the anchor, in cm. Mean $\pm$ sd over seeds.}
\label{tab:tracking-metrics}
\begin{tabular}{@{}lcccc@{}}
\toprule
Method & Seeds & Return & Success & MPJPE (cm) \\
\midrule
\method & 5 & \textbf{0.930}\,$\pm$\,0.004 & \textbf{100.0\%} & \textbf{2.16}\,$\pm$\,0.02 \\
InstantTrack v1.4 & 3 & 0.893\,$\pm$\,0.002 & \textbf{100.0\%} & 2.57\,$\pm$\,0.05 \\
SuperMimic, official code & 3 & 0.819\,$\pm$\,0.057 & 97.5\% & 3.45\,$\pm$\,0.83 \\
\bottomrule
\end{tabular}
\end{table}

\begin{table}[h]
\centering
\small
\caption{Root drift on the 30 LAFAN1 clips longer than 200\,s (225.7--273.1\,s), played once from the first frame on
MuJoCo Warp. None of the three policies observes its root position relative to the reference, only the relative
orientation, so the horizontal pelvis error accumulates. Per seed the mean over the 30 clips, then mean $\pm$ sd over
seeds; end: the error at the end of each clip. Rollouts end only at a fall.}
\label{tab:drift}
\begin{tabular}{@{}lcccc@{}}
\toprule
Method & Seeds & Mean drift (m) & End-of-clip drift (m) & Falls (clips $\times$ seeds) \\
\midrule
\method & 5 & \textbf{0.154}\,$\pm$\,0.018 & \textbf{0.270}\,$\pm$\,0.034 & 0 / 150 \\
InstantTrack v1.4 & 3 & 0.316\,$\pm$\,0.022 & 0.537\,$\pm$\,0.063 & 0 / 90 \\
SuperMimic, official code & 3 & 1.167\,$\pm$\,0.564 & 1.871\,$\pm$\,0.827 & 2 / 90 \\
\bottomrule
\end{tabular}
\end{table}

\section{Additional LAFAN1 results}
\label{app:lafan1}
\label{app:tables}

\begin{figure}[h]
  \centering
  \includegraphics[width=\textwidth]{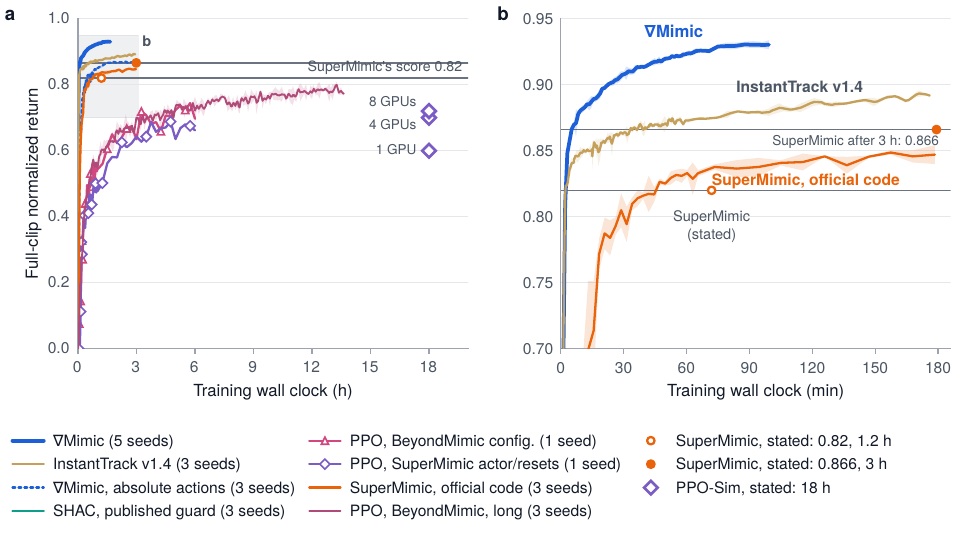}
  \caption{All LAFAN1 runs: full-clip normalized return against training wall clock. (a) Every run over the full
  range. (b) The shaded region of (a), first 186 minutes and returns from 0.70, with the methods closest to \method:
  \method, InstantTrack v1.4, SuperMimic's official code and SuperMimic's stated points. Lines are seed means with a
  $\pm1$ sd band when at least two seeds are evaluated; \method's runs end after about 1.7\,h, once its return has
  levelled off. Every run is scored on MuJoCo Warp. Our runs train on \engine{} on one RTX PRO 6000; InstantTrack and
  SuperMimic's official code train with their own world models in the same physics, and the clocks of SuperMimic's
  official code and of the long PPO runs are converted to the RTX PRO 6000 by update count. SuperMimic (stated) and
  PPO-Sim are the numbers stated in \citet[\S4.1]{liao2026supermimic}.}
  \label{fig:lafan1-main}
\end{figure}

\Cref{fig:lafan1-main} shows every LAFAN1 run against wall clock, including the absolute-action variant of \method,
SHAC, both PPO baselines, the three long PPO runs and SuperMimic's official code, with the numbers stated by
\citet{liao2026supermimic}. \Cref{tab:tab_lafan1_main} lists
all runs, including the H100 and MuJoCo Warp runs and the 6\,h PPO budgets. \Cref{tab:tab_efficiency} gives the cost to
reach 0.82, \cref{tab:tab_ablations,tab:tab_ablations_res} the full ablation results on the absolute- and
residual-action recipes, and \cref{tab:tab_cost} the training-loop cost per policy
update. The earlier main recipe (3 seeds) reaches 0.721 after 3.1\,min, 0.866 after 19.2\,min and 0.890 after
55.1\,min of training, and levels off at $0.914\pm0.004$ without reaching 0.929.

\begin{table*}[t]
\centering
\footnotesize
\caption{LAFAN1: full-clip normalized return at equal training wall clock}
\label{tab:tab_lafan1_main}
\resizebox{\textwidth}{!}{%
\begin{tabular}{lrrrrrrrrrrrrr}
\toprule
Method & Operator & Training hardware & Seeds n & Budget (h) & Return @0.5 h & Return @1 h & Return @2 h & Return @3 h & Return @6 h & Time to 0.82 (h) & Per-seed time to 0.82 (h) & Final return (h) & Failed clips at end (of 40) \\
\midrule
$\nabla$Mimic (NablaEngine) & NablaEngine & RTX PRO 6000 (Blackwell) & 5 & 1.68 & 0.908 $\pm$ 0.002 & 0.923 $\pm$ 0.002 & --- & --- & --- & 0.04 & 0.04 $\pm$ 0.00 & 0.930 $\pm$ 0.004 (1.66 h) & 0.0 $\pm$ 0.0 \\
$\nabla$Mimic, absolute actions (NablaEngine) & NablaEngine & RTX PRO 6000 (Blackwell) & 3 & 3 & 0.829 $\pm$ 0.005 & 0.850 $\pm$ 0.004 & 0.867 $\pm$ 0.001 & 0.870 $\pm$ 0.002 & --- & 0.48 & 0.38 $\pm$ 0.06 & 0.871 $\pm$ 0.003 (3.14 h) & 0.0 $\pm$ 0.0 \\
$\nabla$Mimic, absolute actions (NablaEngine, H100)\textsuperscript{\ddag} & NablaEngine & H100 & 1 & 3 & 0.820 & 0.854 & 0.867 & 0.868 & --- & 0.40 & 0.40 & 0.864 (3.10 h) & 0 \\
$\nabla$Mimic, absolute actions (MuJoCo Warp) & MuJoCo Warp & RTX PRO 6000 (Blackwell) & 3 & 3 & 0.662 $\pm$ 0.016 & 0.743 $\pm$ 0.040 & 0.803 $\pm$ 0.007 & 0.831 $\pm$ 0.009 & --- & 2.39 & 2.42 $\pm$ 0.17 & 0.830 $\pm$ 0.011 (3.00 h) & 0.7 $\pm$ 0.6 \\
SHAC (Xu et al., 2022; published guard), our scale (NablaEngine) & NablaEngine & RTX PRO 6000 (Blackwell) & 3 & 3 & --- & --- & --- & --- & --- & not reached (best 0.388) & --- & 0.384 $\pm$ 0.102 (0.06 h) & 22.7 $\pm$ 4.2 \\
SHAC (published guard), our scale (MuJoCo Warp)\textsuperscript{\ddag} & MuJoCo Warp & RTX PRO 6000 (Blackwell) & 1 & 3 & --- & --- & --- & --- & --- & not reached (best 0.480) & --- & 0.480 (0.36 h) & 17 \\
SHAC (changed guards: non-finite updates skipped, element clip 0.1; NablaEngine) & NablaEngine & RTX PRO 6000 (Blackwell) & 3 & 3 & 0.727 $\pm$ 0.017 & 0.792 $\pm$ 0.010 & 0.810 $\pm$ 0.002 & 0.820 $\pm$ 0.006 & --- & 2.60 & 2.13 $\pm$ 0.34 & 0.820 $\pm$ 0.010 (3.13 h) & 1.7 $\pm$ 1.5 \\
SHAC (changed guards; MuJoCo Warp)\textsuperscript{\ddag} & MuJoCo Warp & RTX PRO 6000 (Blackwell) & 1 & 3 & 0.493 & 0.602 & 0.707 & 0.757 & --- & not reached (best 0.758) & --- & 0.758 (3.13 h) & 5 \\
PPO, BeyondMimic (NablaEngine)\textsuperscript{\ddag} & NablaEngine (forward only) & RTX PRO 6000 (Blackwell) & 1 & 6 & 0.454 & 0.544 & 0.623 & 0.635 & 0.697 & not reached (best 0.744) & --- & 0.697 (6.00 h) & 8 \\
PPO, SuperMimic actor and reset sampling (NablaEngine)\textsuperscript{\ddag} & NablaEngine (forward only) & RTX PRO 6000 (Blackwell) & 1 & 6 & 0.428 & 0.477 & 0.580 & 0.643 & 0.662 & not reached (best 0.688) & --- & 0.662 (6.00 h) & 5 \\
SuperMimic (paper, 1x H100)\textsuperscript{a} & learned world model & H100 (1 GPU) & --- & --- & --- & --- & --- & --- & --- & within 1.2 (stated) & --- & --- & not reported \\
PPO-Sim (SuperMimic paper)\textsuperscript{b} & simulator (paper setting) & H100 (1 / 4 / 8 GPUs) & --- & 18 & --- & --- & --- & --- & --- & not reached in 18 h (stated) & --- & 0.60 / 0.70 / 0.72 (18 h; 1 / 4 / 8 GPUs) & not reported \\
\midrule
\multicolumn{14}{l}{\textit{Appendix rows (earlier recipe)}} \\
$\nabla$Mimic, earlier recipe (v2: 4096 envs, Adam, no normalization) (NablaEngine) & NablaEngine & RTX PRO 6000 (Blackwell) & 3 & 3 & 0.872 $\pm$ 0.006 & 0.892 $\pm$ 0.011 & 0.905 $\pm$ 0.016 & 0.916 $\pm$ 0.003 & --- & 0.12 & 0.11 $\pm$ 0.02 & 0.917 $\pm$ 0.002 (3.07 h) & 0.0 $\pm$ 0.0 \\
$\nabla$Mimic, absolute actions, earlier recipe (no anchor-to-decoder)\textsuperscript{\ddag} & NablaEngine & RTX PRO 6000 (Blackwell) & 1 & 3 & 0.817 & 0.848 & 0.865 & 0.869 & --- & 0.47 & 0.47 & 0.870 (3.13 h) & 0 \\
\bottomrule
\end{tabular}}
\par\smallskip\raggedright\scriptsize
\textsuperscript{\ddag} single seed (no seed variance; a seed-noise reference is given in the notes)\\
\textsuperscript{a} stated: reaches 0.82 within 1.2 h on one H100\\
\textsuperscript{b} stated: does not reach 0.82 within 18 h; final 0.60 / 0.70 / 0.72 on 1 / 4 / 8 GPUs\\
Failed clips = clips with a fall (training termination test) under the nominal evaluation.\\
Falls, $\nabla$Mimic (NablaEngine) (5 seeds): no seed fails a clip at any evaluated checkpoint from update 13,750 (0.95 h) on (the last fall is at update 13,500: 1 failed clip on 1 seed); update 1,250 (0.09 h) is the first checkpoint with 0 of 40 failed clips on all seeds.\\
Falls, $\nabla$Mimic, absolute actions (NablaEngine) (3 seeds): no seed fails a clip at any evaluated checkpoint from update 27,000 (2.88 h) on (the last fall is at update 26,000: 1 failed clip on 1 seed); update 8,250 (0.87 h) is the first checkpoint with 0 of 40 failed clips on all seeds.\\
'(n=k of N)' in a cell: only k of the row's N seeds have a value there (a seed cut at a divergence or not evaluated up to that hour yet); final and crossing columns use only seeds evaluated up to the budget.\\
Training wall clock: training time only for our runs, SHAC and PPO (start-up compilation and all evaluation excluded); SuperMimic and PPO-Sim hours as stated in the SuperMimic paper.\\
SuperMimic's stated numbers are on H100 GPUs; every run of ours and of the baselines we trained is on one RTX PRO 6000 (Blackwell) unless the hardware column says otherwise.\\
Evaluation GPU is A40 or RTX PRO 6000; the actor is deterministic, but MuJoCo Warp scoring is not bit-reproducible: re-scoring one checkpoint varies by up to about 0.004.\\
Seed-noise references (different protocols; neither is a test of a single-seed row): 1-h screen (5 solver iterations), MuJoCo Warp, RTX PRO 6000 (Blackwell): sd 0.0138 of the full return over n=3 seeds; nominal evaluation on NablaEngine runs at matched updates: sd 0.0083 of the full return over n=3 seeds.\\
\end{table*}

\begin{table*}[t]
\centering
\footnotesize
\caption{Cost to reach SuperMimic's LAFAN1 score of 0.82}
\label{tab:tab_efficiency}
\resizebox{\textwidth}{!}{%
\begin{tabular}{lrrrrrrrr}
\toprule
Method & Hardware & n & Time to 0.82 (h) & Policy updates there & Differentiable window steps & Forward-only simulator steps & Total simulator transitions & Per-seed time to 0.82 (h) \\
\midrule
$\nabla$Mimic (NablaEngine) & RTX PRO 6000 (Blackwell) & 5 & 0.04 & 572 updates & 7.49e7 & 5.86e6 & 8.08e7 & 0.04 $\pm$ 0.00 \\
$\nabla$Mimic, absolute actions (NablaEngine) & RTX PRO 6000 (Blackwell) & 3 & 0.48 & 4,548 updates & 1.19e9 & 4.66e7 & 1.24e9 & 0.38 $\pm$ 0.06 \\
$\nabla$Mimic, absolute actions (NablaEngine, H100)\textsuperscript{\ddag} & H100 & 1 & 0.40 & 3,692 updates & 9.68e8 & 3.78e7 & 1.01e9 & 0.40 \\
$\nabla$Mimic, absolute actions (MuJoCo Warp) & RTX PRO 6000 (Blackwell) & 3 & 2.39 & 4,802 updates & 1.26e9 & 4.92e7 & 1.31e9 & 2.42 $\pm$ 0.17 \\
SHAC (Xu et al., 2022; published guard), our scale (NablaEngine) & RTX PRO 6000 (Blackwell) & 3 & not reached (best 0.388 at 0.06 h) & --- & --- & --- & --- & --- \\
SHAC (published guard), our scale (MuJoCo Warp)\textsuperscript{\ddag} & RTX PRO 6000 (Blackwell) & 1 & not reached (best 0.480 at 0.36 h) & --- & --- & --- & --- & --- \\
PPO, BeyondMimic (NablaEngine)\textsuperscript{\ddag} & RTX PRO 6000 (Blackwell) & 1 & not reached (best 0.744 at 5.75 h) & --- & --- & --- & --- & --- \\
PPO, SuperMimic actor and reset sampling (NablaEngine)\textsuperscript{\ddag} & RTX PRO 6000 (Blackwell) & 1 & not reached (best 0.688 at 4.75 h) & --- & --- & --- & --- & --- \\
\midrule
SuperMimic (paper, 1x H100)\textsuperscript{a} & H100 (1 GPU) & --- & within 1.2 (stated) & not stated & --- & not stated & not stated & --- \\
\bottomrule
\end{tabular}}
\par\smallskip\raggedright\scriptsize
\textsuperscript{\ddag} single seed (no seed variance; a seed-noise reference is given in the notes)\\
\textsuperscript{a} stated: reaches 0.82 within 1.2 h on one H100\\
Per-update step counts: $\nabla$Mimic = 4096 x 64 = 262,144 differentiable window steps + 4096 / 0.4 = 10,240 forward-only collector steps; SHAC = 4096 x 32, no collector; PPO = 4096 x 24 (BeyondMimic) or 8192 x 24 (SuperMimic actor and reset sampling) forward steps per iteration.\\
Hours are training time only for our runs (start-up compilation and evaluation excluded); SuperMimic's hours are as stated in its paper.\\
\end{table*}

\begin{table*}[t]
\centering
\footnotesize
\caption{Component ablations of $\nabla$Mimic on the absolute-action recipe (LAFAN1, 3 seeds, \textasciitilde{}1.5 h)}
\label{tab:tab_ablations}
\resizebox{\textwidth}{!}{%
\begin{tabular}{lrrrrrrrr}
\toprule
Change to the final recipe & Component origin & Seeds & Matched at & Return: control & Return: variant & $\Delta$ return & Failed clips (of 40): control & Failed clips (of 40): variant \\
\midrule
No PODS action targets (one Adam step per window) & ours & 3 & 1.64 h (control checkpoints 1.14-1.63 h) & 0.862 $\pm$ 0.002 & 0.809 $\pm$ 0.010 & -0.053 & 0.0 $\pm$ 0.0 & 2.5 $\pm$ 1.0 \\
No per-sample gradient clipping\textsuperscript{a} & ours & 2 of 3 (s0 diverged at 0.54 h, update 5,060) & 1.65 h (control checkpoints 1.18-1.66 h) & 0.862 $\pm$ 0.003 & 0.570 $\pm$ 0.077 (n=2 of 3) & -0.292 & 0.0 $\pm$ 0.0 & 10.4 $\pm$ 5.1 (n=2 of 3) \\
Residual actions instead of absolute setpoints\textsuperscript{b} & action parameterization & 3 & 1.64 h (control checkpoints 1.14-1.63 h) & 0.862 $\pm$ 0.002 & 0.870 $\pm$ 0.001 & +0.008 & 0.0 $\pm$ 0.0 & 0.1 $\pm$ 0.1 \\
Uniform resets instead of return-based reset sampling & inherited: SuperMimic (reset sampling follows BeyondMimic) & 3 & 1.65 h (control checkpoints 1.18-1.66 h) & 0.862 $\pm$ 0.003 & 0.866 $\pm$ 0.001 & +0.004 & 0.0 $\pm$ 0.0 & 0.2 $\pm$ 0.0 \\
No replay starts (windows continue from the previous window's end state) & inherited: SuperMimic (replay-buffer starts) & 3 & 1.62 h (control checkpoints 1.14-1.63 h) & 0.862 $\pm$ 0.002 & 0.879 $\pm$ 0.002 & +0.018 & 0.0 $\pm$ 0.0 & 0.0 $\pm$ 0.0 \\
H=32 windows instead of H=64 & ours & 3 & 1.64 h (control checkpoints 1.14-1.63 h) & 0.862 $\pm$ 0.002 & 0.821 $\pm$ 0.019 & -0.040 & 0.0 $\pm$ 0.0 & 0.4 $\pm$ 0.2 \\
H=32 windows instead of H=64, at matched updates & ours & 3 & update 26,000 (control updates 22,000-26,000, 2.28-2.80 h) & 0.868 $\pm$ 0.002 & 0.821 $\pm$ 0.019 & -0.046 & 0.1 $\pm$ 0.2 & 0.4 $\pm$ 0.2 \\
\bottomrule
\end{tabular}}
\par\smallskip\raggedright\scriptsize
\textsuperscript{a} additional seeds outside the 3-seed arm (not averaged): s3 pending, s4 pending\\
\textsuperscript{b} additional seeds outside the 3-seed arm (not averaged): s3 pending, s4 pending\\
Component origin: 'ours' = introduced by this work; 'inherited' = taken from SuperMimic (whose return-based reset sampling follows BeyondMimic); residual actions change the action parameterization rather than remove a component.\\
Seeds that diverged (non-finite loss) or collapsed (critic value runaway with finite numbers: critic loss above 1e3, return falling to near zero) are listed in the Seeds column and left out of the mean; the variant cell then says '(n=k of 3)'.\\
Seed-noise references (different protocols): 1-h screen (5 solver iterations), MuJoCo Warp, RTX PRO 6000 (Blackwell): sd 0.0138 of the full return over n=3 seeds; nominal evaluation on NablaEngine runs at matched updates: sd 0.0083 of the full return over n=3 seeds.\\
\end{table*}

\begin{table*}[t]
\centering
\footnotesize
\caption{Component ablations of $\nabla$Mimic on the residual-action recipe (LAFAN1, 3 seeds, \textasciitilde{}1.5 h)}
\label{tab:tab_ablations_res}
\resizebox{\textwidth}{!}{%
\begin{tabular}{lrrrrrrrr}
\toprule
Change to the final recipe & Component origin & Seeds & Matched at & Return: control & Return: variant & $\Delta$ return & Failed clips (of 40): control & Failed clips (of 40): variant \\
\midrule
No PODS action targets (one Adam step per window) & ours & 3 & 1.63 h (control checkpoints 1.14-1.58 h) & 0.871 $\pm$ 0.001 & 0.830 $\pm$ 0.015 & -0.041 & 0.1 $\pm$ 0.1 & 1.3 $\pm$ 0.8 \\
No per-sample gradient clipping & ours & 0 of 3 (s0 diverged at 0.59 h, update 5,510; s1 diverged at 0.31 h, update 2,930; s2 diverged at 0.37 h, update 3,500) & --- & --- & --- & --- & --- & --- \\
Uniform resets instead of return-based reset sampling\textsuperscript{a} & inherited: SuperMimic (reset sampling follows BeyondMimic) & 3 & 1.64 h (control checkpoints 1.14-1.58 h) & 0.871 $\pm$ 0.001 & 0.813 $\pm$ 0.062 & -0.058 & 0.1 $\pm$ 0.1 & 3.1 $\pm$ 2.9 \\
No replay starts (windows continue from the previous window's end state) & inherited: SuperMimic (replay-buffer starts) & 3 & 1.64 h (control checkpoints 1.14-1.58 h) & 0.871 $\pm$ 0.001 & 0.891 $\pm$ 0.006 & +0.020 & 0.1 $\pm$ 0.1 & 0.0 $\pm$ 0.0 \\
H=32 windows instead of H=64\textsuperscript{b} & ours & 3 of 4 (s2 collapsed (critic value runaway) at 1.10 h, update 17,080) & 1.65 h (control checkpoints 1.15-1.66 h) & 0.871 $\pm$ 0.001 & 0.831 $\pm$ 0.009 (n=3 of 4) & -0.040 & 0.1 $\pm$ 0.1 & 0.3 $\pm$ 0.2 (n=3 of 4) \\
H=32 windows instead of H=64, at matched updates\textsuperscript{c} & ours & 3 of 4 (s2 collapsed (critic value runaway) at 1.10 h, update 17,080) & update 25,583 (control updates 21,000-25,000, 2.18-2.63 h) & 0.878 $\pm$ 0.003 & 0.831 $\pm$ 0.009 (n=3 of 4) & -0.046 & 0.0 $\pm$ 0.0 & 0.3 $\pm$ 0.2 (n=3 of 4) \\
\bottomrule
\end{tabular}}
\par\smallskip\raggedright\scriptsize
\textsuperscript{a} a transient collapse falls inside the averaging window; the last checkpoints of all three seeds are 0.86-0.87\\
\textsuperscript{b} additional seeds outside the 3-seed arm (not averaged): s3 completed\\
\textsuperscript{c} additional seeds outside the 3-seed arm (not averaged): s3 completed\\
Component origin: 'ours' = introduced by this work; 'inherited' = taken from SuperMimic (whose return-based reset sampling follows BeyondMimic); residual actions change the action parameterization rather than remove a component.\\
Seeds that diverged (non-finite loss) or collapsed (critic value runaway with finite numbers: critic loss above 1e3, return falling to near zero) are listed in the Seeds column and left out of the mean; the variant cell then says '(n=k of 3)'.\\
Seed-noise references (different protocols): 1-h screen (5 solver iterations), MuJoCo Warp, RTX PRO 6000 (Blackwell): sd 0.0138 of the full return over n=3 seeds; nominal evaluation on NablaEngine runs at matched updates: sd 0.0083 of the full return over n=3 seeds.\\
\end{table*}

\begin{table*}[t]
\centering
\footnotesize
\caption{Training-loop cost per policy update}
\label{tab:tab_cost}
\resizebox{\textwidth}{!}{%
\begin{tabular}{lrrrrrrrr}
\toprule
Method & Operator & Hardware & Seconds per policy update (median) & Policy updates per hour (wall-clock slope) & Differentiable steps per update & Forward-only steps per update & Simulated steps per second & Engine speed-up vs MuJoCo Warp \\
\midrule
$\nabla$Mimic\textsuperscript{a} & NablaEngine & RTX PRO 6000 (Blackwell) & 0.240 & 14,384 & 131,072 & 10,240 & 588k & 6.8x \\
$\nabla$Mimic, absolute actions\textsuperscript{b} & NablaEngine & RTX PRO 6000 (Blackwell) & 0.373 & 9,383 & 262,144 & 10,240 & 729k & 4.4x \\
$\nabla$Mimic, absolute actions, hoisted encoder (timing run)\textsuperscript{c} & NablaEngine & RTX PRO 6000 (Blackwell) & 0.369 & --- & 262,144 & 10,240 & 738k & 4.4x \\
$\nabla$Mimic, absolute actions & MuJoCo Warp & RTX PRO 6000 (Blackwell) & 1.633 & 1,966 & 262,144 & 10,240 & 167k & --- \\
\midrule
$\nabla$Mimic, absolute actions (H100) & NablaEngine & H100 & 0.382 & 9,275 & 262,144 & 10,240 & 712k & 5.9x\textsuperscript{\S} \\
$\nabla$Mimic, absolute actions (H100)\textsuperscript{d} & MuJoCo Warp & H100 & 2.258 & --- & 262,144 & 10,240 & 121k & --- \\
\midrule
SHAC, our scale & NablaEngine & RTX PRO 6000 (Blackwell) & 0.196 & 18,365 & 131,072 & 0 & 669k & 4.3x \\
SHAC, our scale & MuJoCo Warp & RTX PRO 6000 (Blackwell) & 0.839 & 3,455 & 131,072 & 0 & 156k & --- \\
\midrule
PPO, BeyondMimic\textsuperscript{e} & NablaEngine (forward only) & RTX PRO 6000 (Blackwell) & 0.126 & 28,240 & 0 & 98,304 & 781k & --- \\
PPO, SuperMimic actor and reset sampling\textsuperscript{f} & NablaEngine (forward only) & RTX PRO 6000 (Blackwell) & 0.549 & 6,542 & 0 & 196,608 & 358k & --- \\
\bottomrule
\end{tabular}}
\par\smallskip\raggedright\scriptsize
\textsuperscript{a} residual actions\\
\textsuperscript{b} final recipe\\
\textsuperscript{c} speed reference for the final recipe's encoder hoisting (bitwise-equivalent up to matmul rounding); 70-update timing run\\
\textsuperscript{d} O64 recipe without anchor-to-decoder: a different recipe from the final engine runs; 30-update timing run\\
\textsuperscript{e} per PPO iteration\\
\textsuperscript{f} per PPO iteration\\
Simulated steps per second count differentiable window steps (forward + adjoint) and forward-only steps together.\\
The engine runs of the paper pin one engine release; earlier timing runs used the engine build of their day, so per-update times across rows mix engine versions with recipe changes.\\
\textsuperscript{\S} on a speed-up: the MuJoCo Warp timing comes from a run of a different recipe (named in its notes), so the speed-up mixes the operator with recipe changes; the matched-recipe figure needs the final MuJoCo Warp run.\\
\end{table*}

\end{document}